\RequirePackage{fix-cm}

\documentclass[Afour,sageh,times]{sagej}

\newcommand\BibTeX{{\rmfamily B\kern-.05em \textsc{i\kern-.025em b}\kern-.08em
		T\kern-.1667em\lower.7ex\hbox{E}\kern-.125emX}}

\usepackage{epstopdf}
\usepackage{graphicx}
\usepackage{epsf}
\usepackage{amsmath}
\usepackage{amsfonts}
\usepackage{algorithm}
\usepackage[noend]{algorithmic}
\usepackage{algorithmic}
\usepackage{color}
\usepackage{rotating}
\usepackage{subfigure}
\usepackage{moreverb,url}

\usepackage{totcount}

\newtotcounter{citnum} 
\def\oldbibitem{} \let\oldbibitem=\bibitem
\def\bibitem{\stepcounter{citnum}\oldbibitem}

\usepackage[colorlinks,bookmarksopen,bookmarksnumbered,citecolor=red,urlcolor=red]{hyperref}

\newcommand{\sbf}[1]{\boldsymbol{#1}}

\begin{document}

	\date{Received: date / Accepted: date}

	\runninghead{Jay M. Wong --- Towards Lifelong Self-Supervision: A Deep Learning Direction for Robotics}
	
	\title{Towards Lifelong Self-Supervision:\\ A Deep Learning Direction for Robotics}
	
	\author{Jay M. Wong} 
	
	\affiliation{
		The Charles Stark Draper Laboratory, Inc,  Cambridge, MA, USA}
	
	\corrauth{Jay M. Wong, Planning, Autonomy, and Automation Group, The Charles Stark Draper Laboratory, Inc, 555 Technology Square, Cambridge, MA, USA}
	
	\email{\texttt{jmwong@draper.com} \\
		\mbox{}\\\textbf{This document contains \total{citnum}\ references.} 
	}

	\keywords{Robotics, Deep Learning, Cognition, Autonomy, Lifelong Learning}

	\begin{abstract}
		Despite outstanding success in vision amongst other domains, many of the recent deep learning approaches have evident drawbacks for robots. This manuscript surveys recent work in the literature that pertain to applying deep learning systems to the robotics domain, either as means of estimation or as a tool to resolve motor commands directly from raw percepts. These recent advances are only a piece to the puzzle.  We suggest that deep learning as a tool alone is insufficient in building a unified framework to acquire general intelligence. For this reason, we complement our survey with insights from cognitive development and refer to ideas from classical control theory, producing an integrated direction for a  lifelong learning architecture. 
	\end{abstract}
	
	\maketitle
	
	\renewcommand*{\thefootnote}{\arabic{footnote}}
	\setcounter{footnote}{0}
	
	\section{Introduction}
	As roboticists and scientists, is our goal to engineer a solution to work for a particular task in a specified domain or is it to build a system that has the capacity to acquire general intelligence (a notion characterized by \cite{Legg2007})? An historic example is that success in aviation, with autonomous aerial navigation does not immediately imply success in a task like autonomous driving, which shares some similarities. Or assume that we solve autonomous driving tomorrow---likely an engineering effort like changing our highways, roads, and infrastructure with  increased sensory \cite{Ng2016} will not generalize especially well to robotic tasks in mobile manipulation. Should we then add additional sensors to all possible environments (e.g. residential homes) where autonomous systems are likely to operating in? How is it then to operate in novel, unknown, or disastrous environments? Or in space? In fact, it is under inspirations from these disastrous and unstructured domains, that have given rise to recent technological advances with the DARPA Robotics Challenge (e.g. \cite{Johnson2015,Feng2015,Feng2015b,Kohlbrecher2015,Yi2015,Kuindersma2016,Dellin2016}). This manuscript presents rather a different direction in thinking, where instead of engineering and redesigning systems to perform competently in novel tasks and domains, perhaps a system that can bootstrap its lifetime of experiences can quickly learn useful solutions in these new areas---we refer to the process by which this long-term knowledge repertoire is acquired as \emph{lifelong learning}.
	
	Lifelong learning should not address only novel domains, but also should consider optimizing behavior at existing tasks. Let's consider the following hypothetical scenario. An autonomous system (e.g. robot, mobile manipulator, unmanned aerial vehicle (UAV), etc.) returns from a mission or accomplishes some task. We are now out of things to provide it. Likely in many cases, the robot is left somewhere in corner of a laboratory until there are subsequent tasks to accomplish. But what if instead, it uses this downtime as an emergent possibility for continuous progress?---and continue to operate, either refining its inherent representations of the world (which have generally, to this date, been hand-defined by human operators) or optimizing its inherent motion primitives (e.g. tuning internal control parameters that perhaps due to wear and tear are now highly suboptimal). What if it can learn to build complex motor behaviors, that may prove useful in future missions from exploiting existing structure in its primitives? 
	
	A misconception is that what we are referring to as lifelong learning does not necessarily imply that the system learns from scratch. It is \emph{not} an end-to-end approach for motor development or task solving. In other words, it does not necessarily imply learning \emph{motor torques} directly from raw sensory input. Instead, its underlying purpose is generalization and \emph{structural bootstrapping}, a term coined by \cite{Worgotter2015}, where existing knowledge is exploited for generalization to novel activities. We draw insight from cognitive development to learn complex motor behavior and structural representations by an account of intrinsic and extrinsic properties (e.g. environmental uncertainties) influencing the system in ways beyond engineering analyses. Lifelong learning implies that systems learn over a lifetime of complex tasks and domains, achieving generalized solutions. This form of generalizability for both domain and task is extremely important for designing robust, high-performance systems. Interestingly, a study by \cite{Pinto2016} found that convolutional networks achieved higher performance grasping when trained on both grasping and pushing tasks when compared to grasping alone, suggesting that inter-task representation sharing helps build a better understanding of the environment overall. 
	
	This manuscript provides an initial survey on recent advances in deep learning  pertaining to the robotics domain and complements this review with inspirations in cognitive development and control theory outlining that the coalescence of these ideas may pave way for lifelong learning robots. We indicate that deep learning alone is likely incapable of solving all problems in a unified framework. Instead, we discuss a connectionist approach in lifelong self-supervision, drawing ideas from these other areas to tackle the acquisition of general motor intelligence. We formulate a direction by discussing how systems can intrinsically motivate themselves to attack the problem of building accurate representations of structures in the world \emph{and} the development of complex motor behavior \emph{simultaneously}. This particular direction incorporates the use of hierarchies of neural networks under the popularized notion of \emph{deep learning}. We suggest that the use of deep learning as an approximation tool allows robots to encode complex functions that describe physical phenomena concerning interactions with the world and sensory-driven control. 
	
	\subsection{From Computer Vision to Robotics} Deep learning has exhibited major success stories in the computer vision domain. This particular tool, popularized by \cite{Krizhevsky2012} in the ImageNet competition, showed significant promise when learned latent feature representations by a neural network that incorporated a series of convolution, pooling, and densely connected neurons outperformed existing hand crafted feature representations that have otherwise been the standard. The support from computational machinery (GPUs) allowed for efficient parallelization necessary for training neural network structures with massive datasets. Since then, the computer vision community has produced a plethora of deep learning research and have plateaued close to human level capabilities in recognition by building deeper and deeper networks \cite{Szegedy2015}, fine tuning, and introducing extra features (e.g. surface normals) \cite{Madai2016}. However, that is not to say that these powerful, state of the art, demonstrations and their solutions are immediately applicable to mobile perceptual systems. There exists a number of fundamental differences in these two domains that hinder trivial compatibilities. First, the deep learning solutions in computer vision are generally supervised. Supervised learning is very constrained. And in many computer vision tasks, a particular input is associated with a single, correct output. 
	
	Yet quite evidently, this is not the case in robotics---robots do more than classification. They must perform actions in the world. They must build representations of things they sense and act on these sensory signals whereas computer vision systems do not necessarily act. 
	Classification helps in identifying the entities in the world, but to accomplish tasks, robots must perform actions and manipulate such entities. The connection between perception and action is essential in building perceptual systems in the real world. 
	
	\subsection{On Overgeneralization} An immediate drawback for these computer vision architectures is that studies have found that image classification has an unfortunate overgeneralization (``fooling'')  phenomenon. These classification tasks take as input generally a single sensor modality, in many cases, RGB. Where deep learning tools fail is when adversarial RGB examples are construed in attempts to fool these networks into very incorrect predictions presented in studies by \cite{Szegedy2013,Nguyen2015,Carlini2016}. In these works, hill climbing and gradient ascent methods were used to evolve images to match very incorrect classes with high probability predicted by the network. Although there is work to make these networks robust to such malicious attacks (e.g. \cite{Bendale2015,Wang2016b})\footnote{This particular phenomenon is categorized under \emph{open set recognition}. \cite{Bendale2015} proposed a new model layer, OpenMax, that estimates the probability of input from unknown classes and rejects fooling adversarial examples.}, we theorize that the addition of multiple modalities (with more than vision alone) may alleviate such intriguing and devastating phenomena, as the real world obeys certain structures that are locally constraining \cite{Kurakin2016}. For instance, it is increasingly difficult to fool a predictor that reasons with depth and tactile information with physical adversarial entities. Furthermore, by the universal approximator theorem \cite{Hornik1989}, multilayer feedforward networks are capable of representing arbitrarily complex functions, even those that are robust to such adversarial anomalies. It becomes then a formulation problem where models must be trained with an adversarial objective \cite{Goodfellow2015a}. Other work elaborates that networks should be also realized with proper regularization \cite{Tanay2016}. In fact, a particular variant of neural networks that implicitly enforces regularization may be beneficial. For instance, \emph{DivNet} is an approach that attempts to model neuronal diversity allowing for efficient auto-pruning, in turn reducing network size and providing inherent regularization \cite{Mariet2015}. Another remedy realized a particular network layer called \emph{competitive overcomplete output layer} to mitigate this overgeneralization problem of neural networks \cite{Kardan2016}. Such a layer forces outputs to explicitly compete with each other, resulting in tight-fitting regions around the training data. 
	
	\subsection{On Deep Reinforcement Learning} Impressive success stories has been shown in game domains that integrate perception and action. Namely work by Google DeepMind has pushed this particular frontier and have revolutionized the intersection between deep learning and its connection to reinforcement learning. \cite{Mnih2013,Mnih2015} built a single learning framework that could learn to play a large number of Atari games beyond human level competence through a trial and error approach, where the only information given to the system were several game frames, the game score, and a discrete control set. A neural network was used to approximate the $Q$ values of a discrete set of actions from several game play frames and executed the highest valued action resulting in a competent gameplay policy. The idea of using a neural network as an approximation to a value function is not something profound and novel. Dating back two and a half decades ago, Tesauro attempted to tackle the game Backgammon---a board game that had approximately $10^{20}$ states, making traditional table-based reinforcement learning infeasible. Instead, a backpropagation layered neural network was used to approximate the value function describing board positions and probabilities of victory. It was shown in various incarnations of his algorithm where both raw encodings and hand-crafted features derived from human task knowledge were used to learn competent gameplay policies \cite{Tesauro1992,Tesauro1995,Tesauro1995b}. Furthermore, other researchers in the past have leveraged recurrent neural networks to learn $Q$ values using a history of features to form policies \cite{Jurgen1991,Lin1992,Meeden1993}.
	
	Recently, the defeat of 18-time Go world champion, Lee Sedol, by DeepMind's AlphaGo system established a major milestone for deep learning frameworks. Their success was not simply attributed to deep reinforcement learning but also clever integration with Monte Carlo tree search \cite{Silver2016}---it is to show that deep learning alone may not be the solution to all problems, but as a tool, deep learning may used in complement with other algorithms to produce very powerful results. 
	
	\subsection{On Domain Transfer} Unfortunately, a reason why many of these game domains are successful is that the domain is fully observable. Despite there being some studies that inject partial observability and attempt to tackle this problem with augmented memory structure \cite{Heess2015}, the algorithms developed through gameplay should not be considered as game-changing success stories in physical dynamical systems. The real world obeys physics and uncertainties that can not be perfectly modeled in simulation, and as a result policies learned in simulation have difficulty generalizing to real robot systems. For instance, despite millions of training steps in over hundreds of hours of simulations, visuomotor policies learned on a Baxter simulator fails when given real world observations on the physical platform \cite{Zhang2015}. Interestingly, \cite{James2016} were able to transfer visuomotor policies learned in simulations to a real system, however, their approach required massaging the scene by occluding complex areas with a physical black box, hiding wires, and mimicking the simulation setting. The introduction of \emph{progressive neural networks} show promise by exploiting deep composition of features amongst columns of domain-specific networks. However, immediate drawbacks are that it assumes known task boundaries and exhibits quadratic parameter growth when a new column is necessary for each novel domain. Despite alleviation to the cause, these networks still need to be trained in the new domain to achieve competence \cite{Rusu2016}. In later works, \cite{Rusu2016b}, demonstrated domain transfer  using features learning in MuJoCo simulation with a Kinova arm to the physical system---the reaching task they showed, however, was highly constrained in a small region of static space and still require several hours of training in the real world. A generalized method for domain alignment has recently been presented to mitigate performance loss when adapting robot visuomotor representations from synthetic to real environments \cite{Tzeng2015}. The technique, however, requires paired synthetic and real views of the same scene to adapt the deep visual representations.  
	
	\subsection{On Self-Supervision} As a result, there is no escaping the fact that robots must collect their own training data in the real world---to tackle this, various approaches by \cite{Pinto2015,Levine2016,Wong-RSS2016} have been proposed for self-supervision. These studies hint at methods in which robots can label their own experiences and collect potentially massive datasets pertaining to a single task. However, they provide no notion of learning beyond the task at hand---conceptually they are incapable of exhibiting lifelong learning. To address this, we suggest a direction in which robots acquire completely unsupervised visuomotor skills derived from a basis of inherent primitives that are reinforced by the world to generate behaviors that adhere to physical properties of the environment. This hierarchical set of motor behaviors should then be coupled with an intrinsically motivated structure learning module to allow continuous affordance and interaction outcome prediction regarding entities in the world---as a result, this produces permanent artifacts that can be reused for future tasks. Furthermore, the intrinsic motivator must both promise the acquisition of continuously refined forward models and the capacity for the development of arbitrarily complex motor behaviors. 
	
	This paper is in agreement with \cite{Silver2013} in which they address the machine learning community that we need to seriously consider the nature of systems that have capacity to learn over a lifetime of experiences rather than for some specified task or domain. Conceptually, this direction of thinking is relatable to the notion of \emph{deep developmental learning} primarily proposed by \cite{Sigaud2016} in which first sensory motor control must transform raw sensations to a predictive process---we refer to this as a forward affordance model. They also outlined the challenges of integrating behavioral optimization and a curiosity mechanism for deep developmental systems. We suggest to attack these simultaneously through the continuously refinement of motion primitives and the exploitation of their combinatoric sequences and compositions to achieve motor development. To address the latter, we suggest instrinsic motivators driven by information theoretic measures in regards to the forward affordance model's predictions of world state. Lastly, we quickly address a \emph{certain debate} dating back two decades between planning and reflexive architectures for the design of robot behavior \cite{Brooks-FAI87,Brooks1987}. While we agree that hierarchical behavioral responses eliminates the need for planning, we take a stand that is much similar to our ideologies with deep learning---that is, to reiterate, a grandiloquent singular architecture is likely infeasible in many senses. Behavioral responses need not be learned when set rules and instructions are provided (e.g. autonomous missions and manuals where there are precise trajectories through the task)---perhaps this is where we should consider planning for task solving. As such, we firmly believe that there is no single individual or rather, what Rodney Brooks calls \emph{theists}, that is truly correct. Instead, we find that excitement is in building a system that marries the many \emph{theisms} and relates technologies that are both classical and of recent ``hype.'' As such, this leads us to a unifying, perhaps even holistic, direction in thinking. To our knowledge, this paper is the first attempt at outlining a lifelong learning direction by integrating deep learning, cognitive development, and classical control theory.

	\section{Implications for Robotics} 
	In the most simplifying sense, deep learning is an algorithmic tool that leverages neural networks as nonlinear function approximators in which weighted connections between input and output neurons are trained via error back propagation. Doing so, encodes a function that minimizes the disparity between prediction and truth by building latent representations in the hidden layers. The deep learning domain has been quickly exploding since its large success in vision popularized by the outstanding results in the ImageNet competition \cite{Krizhevsky2012}, producing a plethora of general reviews on deep neural networks. As a result, we omit the basics of neural networks, convolutions, autoencoding, regularization, recurrency, and related concepts in this manuscript. The reader is referred to the following general reviews \cite{Bengio2009,LeCun2015,Goodfellow2016} for a thorough overview. Instead, we will critique a number of deep learning frameworks most applicable to the \emph{robotics} domains and outline the drawbacks and skepticism that arise with these recent works. 
	
	\subsection{Detection,  Estimation, and Tracking}
	
	Tools that have been popularized by the computer vision community has generally been leveraged to tackle individual components in the robotics domain. A number of these individual triumphs used neural networks as a means of approximating otherwise very complex and highly nonlinear functions pertaining to estimation and scene understanding. 
	
	For instance, tasks relating to rule-based navigation like autonomous driving in particular may require understanding the identities of objects in the world. Labeling and scene understanding can be regarded as a segmentation problem given visual input. As such, SegNet, a semantic pixel-wise segmentation encoder-decoder network, was presented to achieve competitive predictive capability \cite{Badrinarayanan2015}. Other methods also attempted to solve a similar task but were either generating object proposals \cite{Noh2015,Hariharan2015} or required multi-stage training \cite{Socher2011,Zheng2015}. Still, these segementation results were shown to be especially robust for detection problems. In particular, \cite{Pinheiro2015} built a system to predict segmentation masks given input patches by using \emph{DeepMask}, a neural network that generated such proposals. These proposals were then passed to an object classifier, producing state of the art segmentation results. Extensions to this work gave way to a bottom-up/top-down segmentation refinement network that was capable of generating high fidelity object masks with a $50\%$ speedup. The network, \emph{SharpMask} leveraged features from all layers of the network by first generating a course mask prediction and refining this mask in a top-down fashion \cite{Pinheiro2016}.
	
	Studies have shown that pre-trained convolutional neural network features were useful for RGB-D object recognition and pose estimation \cite{Schwarz2015}. A consequence of this became a flurry of research regarding using deep architectures for detection and pose estimation. An approach for the detection of pedestrians was shown using an unsupervised multi-stage feature learning approach by \cite{Sermanet2013}. Meanwhile, results indicated high accuracy in human pose estimation with \emph{DeepPose}---likely this is due to deep neural networks capturing context and reasoning in a holistic manner \cite{Toshev2014}. 
	
	PoseNet, a convolutional neural network camera pose regressor, was presented with impressive robustness to difficult lighting, motion blur, and different camera intrinsics \cite{Kendall2015}. This was later extended to a Bayesian model able to provide localization uncertainty \cite{Kendall2015b}, establishing a critical step forward for mobile robots, especially connecting close ties to algorithms that operate under uncertainty. In work by \cite{Wilkinson2015}, a pretrained convolutional network was used to predict object descriptions and aspect definitions pertaining to sensory geometries in relation to objects. Unfortunately, class and object descriptors were selected as arbitrary pretrained AlexNet layers and the overall framework relied on a number of thresholds that are difficult to define. 
	
	Others continue to investigate the use of these tools to learn useful features for contexts like laser based odometry estimation \cite{Nicolai-RSS2016}. In research by \cite{Byravan2016}, deep networks were used to segment rigid bodies in the scene and predict motions of these entities in SE3.
	
	As hyperparametric approximators, these networks have been found success in the tracking regime in which recurrent neural networks were used to filter raw laser measurements and shown to infer object location and identify in both visible and occluded scenes \cite{Ondruska2016}. This technique is described as a neural network analogous to Bayesian filtering. In addition, by learning to track with a large set of unsupervised data, a new task like semantic classification could be learned by exploiting rich internal structure through inductive transfer \cite{Ondruska-RSS2016}. An approach was presented by \cite{Song2015} where 3D bounding boxes of objects were generated through a methodology they call \emph{Deep Sliding Shapes}. Given Kinect images, they learned a multiscale 3D region proposal network that is fully convolutional and identifies interesting regions in the scene. Then, an object recognition network was learned to perform 3D box regressions.

	A sensory-fusion architecture that incorporated the use of LSTMs to capture temporal dependencies has been presented to anticipate and fuse information from multiple sensory modalities. This Fusion RNN was demonstrated as part of a maneuver anticipation pipeline that outperformed state of the art on a benchmark consisting of a dataset of $1180$ miles of natural driving \cite{Jain2016}. Similarly, \cite{Krishnan2015} developed a deep network capable of approximating a broad class of Kalman filters, enhancing them to arbitrarily complex transition dynamics and emission distributions.
	
	Yet, despite outstanding results in classification, identification and pose estimation, and semantic segmentation, systems that perform actions in the world still require a connection from detected entities in space to motor commands. In part, these research solutions only attempt to develop robust perceptual interfaces to autonomous systems, but however, a key, perhaps, paramount module is one that reasons over sensations and executes useful motor control. As such, perception alone may not be the answer, but somewhere in the intersection of perception, cognition, and action.

	\subsection{From Perception to Motor Control}
	A number of studies looked into introducing the predictive power of neural networks in place of traditional feature extracting perceptual pipelines to solve detection and control problems with physical robot experiments. In particular, in place of hand-designed features like those of \cite{Kragic2003,Maitin2010} for grasping, \cite{Lenz2015} presented a deep architecture to learn useful feature representations for grasp detection. A two-step cascaded network system was shown where top detections were re-evaluated by the second network, allowing for quick pruning of unlikely candidate grasps. The network operated on RGB-D input and successfully generalized to execute grasps on both a Baxter and PR2 robot. Likewise, a grasp detection system was demonstrated by \cite{Wang2016}, that mapped RGB-D images to gripper grasping pose by first segmenting the graspable objects from the scene using geometric features (for both objects and gripper). They then applied a convolutional network to the graspable objects which used a structure penalty term to optimize the connections between modalities. Similarly, in work presented by \cite{vanHoof2016}, deep autoencoders were used to learn compact latent representations for reinforcement learning to form policies describing tactile skills. These feedback policies were learned directly from high-dimensional space under iterative on-policy exploration and vastly outperformed a baseline policy learned directly from the raw sensor data. 
	
	Contrary to these works, instead of predicting the single best grasp pose from a given image, \cite{Johns2016} demonstrated a convolutional network that predicted a score for every possible grasp pose, such a value function described what they denote as a \emph{grasp function}. They discussed that such a method can attribute to robust grasping by smoothing this grasp function with a function describing pose uncertainty. Although in their demonstrations, it appears this particular approach achieved some-$80\%$ grasp success rate, fundamental assumptions are that the object is isolated in the scene and the grasping device is a parallel jaw gripper. 
	
	In a different approach, \cite{Varley2016} showed that convolutional networks can be used for shape completion given an observed point cloud. In their method, the network learned to predict a complete mesh model of objects (filling in the occluded regions of the scene), which was then smoothed and used to support grasp planning. 
	
	The use of convolutional neural networks as a means for automatic feature extraction has been employed in imitation learning paradigms where actions are learned for an autonomous navigation task directly from raw visual data. The network is encoded with no initial knowledge of the task, targets, or environment in which it is acting in. In a simulated study \cite{Hussein2016} showed that using \emph{deep active learning} can significantly improved the imitated policy through a small number of samples---this is accomplished by the network querying a teacher for the correct action to take in situations of low confidence. Unfortunately, this framework relies on the fact that there is a teacher present with competent knowledge of the domain and appropriate actions. A framework using  time-delay deep neural network was shown by \cite{Noda2013} that both fused multimodal sensory information and learned sensorimotor behaviors simultaneously. They demonstrated that a single network was able to encode six object manipulation behaviors dependent on temporal sequence changes with the environment and displayed object. 
	
	
	Since robots operate in dynamic and partially observable environments, selecting the best action is nontrivial since it is dependent on the time history of interactions (or sequences of actions in the past). As such, ways to learn these optimal policies generally rely on a trial and error paradigm via reinforcement learning. For example, this is especially present in the recent successful demonstrations of Atari gameplay \cite{Mnih2013,Mnih2015}. Despite its demonstration through an artificial agent, the Deep $Q$ Networks presented has immediate implications in robot control, however, may not be effective on physical domains especially when rewards are sparse making efficient exploration essential. A method proposed by \cite{Lipton2016} demonstrated exploration by Thompson sampling where using Monte Carlo samples from a Bayes-by-Backprop neural network provided improvement over the standard DQN approach that relied either on $\epsilon$-greedy or Boltzmann exploration. 
	
	In a particular study, \cite{Finn2016a} proposed to use neural networks as a tool to learn arbitrarily complex and nonlinear cost functions for inverse optimal control problems allowing systems to learn from demonstration using efficient sample-based approximations. Their methods were demonstrated on simulated tasks as well as on a mobile manipulator. Another study presented a belief-driven active object recognition system that used a pretrained AlexNet first to derive belief state. A Deep $Q$ Network was then incorporated to actively examine objects by selecting actions (in hand manipulations) that minimized overall classification errors, resulting in an efficient policy for recognizing objects with high levels of accuracy \cite{Malmir2016}. Instead of training the action selection network over the pretrained convolutional network, this system was later extended to be trainable end-to-end \cite{Malmir2015}. 
	
	In contrast to the large population of work that uses convolution to extract useful feature representations at the output layer of a neural network and use these features to associate control, \cite{Ku2016} demonstrated a different approach. They showed that using intermediate features of a convolutional network was sufficient for gross and finer grain manipulation supporting palm and finger grasps. The technique localized features corresponding to high activations given point clouds of simple household objects through \emph{targeted backpropagation}. Using this, they presented a hierarchical controller composing of finger and palm pre-posture positions on the R2 robot, however, alike work by \cite{Wilkinson2015}, the specific layer to obtain information from is still human defined.

	\subsubsection*{From Pixel to Motion --} Recently, an end-to-end strategy for visuomotor control popularized by \cite{Levine2015} has shown promise for deep learning in robotics. Using optimal control policies as supervised signal for neural networks, they demonstrated task learning relevant to local spatial features obtained through convolution. More importantly, this showed promise for an end-to-end training approach to obtain visuomotor policies producing a network that commanded motor torques directly from the raw visual input. \cite{Finn2015} proposed the use of deep spatial autoencoders to acquire informative feature points that correspond to task-relevant positions. The method learns to associate motions with these points using an efficient locally-linear reinforcement learning method---because the resulting policies are based off these learned feature points, the robot is capable of dynamically manipulation in a closed-loop manner. Similar approaches to visuomotor control has been demonstrated by \cite{Tai2016} the learned end-to-end exploration policies on a mobile robot and folks from nVidia for self-driving cars, where an autonomous vehicle was driven by vision alone through an end-to-end system \cite{Bojarski2016}. It appears that learning motor control directly from raw sensory signals induces robustness and produces a control solution that is otherwise too complex to hand design. Likely, action outcomes are not deterministic and pose estimation future establishes uncertainties, whereas, these convolutional learning strategies aims to resolve motor commands straight from raw percepts---learning both useful feature representations and control policies simultaneously. 
	
	An issue with end-to-end methods and deep reinforcement learning in general is that it demands extremely large sets of data. For such reasons, Guided Policy Search (GPS) \cite{Levine2015} attempts to bias training for the reduction of the number of instances needed and looked to acquire visuomotor policies by the means of a supervised learning problem. A reset-free GPS algorithm was introduced by \cite{Montgomery2016b} to address the issue with its requirement for a consistent set of initial states. Meanwhile, \cite{Chebotar2016} later extended GPS to account for highly discontinuous contact dynamics through an path integral optimizer and on-policy sampling to increase the diversity of instances which they argued was crucial for high generalizability.
	
	In the original formulation of GPS, the learning problem is decomposed into a number of stages. First full-state information is used to create locally-linear approximations to the dynamics around nominal trajectories, then optimal control is used to find locally-linear policies along those trajectories. Lastly, it uses supervised learning with an Euclidean loss objective to create complex nonlinear versions of these policies that reproduce similar optimized trajectories. In other words, GPS iteratively optimizes local policies (concerning specific task instances) which are then used to train a global policy that is general across instances. However, to do this, it requires data on the physical system---robots must collect data for training to refine the network originally trained by guided policies in the form of optimal trajectories. 
	
	To collect massive sets of data, one may consider having the robot obtain its own experiences without the need of meticulous human labeling or supervision. Addressing the problem of self-supervision, work by \cite{Pinto2015} showed that robots can self-supervise themselves to learn visuomotor skills without manual labels. In their experiments, they demonstrated remarkable robustness where a Baxter robot self-labeled $50,000$ grasp examples in over $700$ hours of manipulation. Under similar inspirations, \cite{Levine2016} used a distributed system consisting of $14$ robot manipulators to collect a massive dataset ($800,000$ grasps) over the course of two months for grasping and eye-hand coordination. However, both of these self-supervised methods considered specifying a heuristic to classify grasp examples. Unfortunately, these heuristics are human-specific and somewhat arbitrary. \cite{Wong-RSS2016} developed a comparable self-supervision approach with a key distinction being the use of feedback from closed-loop motion primitives as a supervisory signal rather than these human-specific parameters. Still, a fundamental problem for all of these studies is that they are demonstrated and tailored for a single specific, predefined task. A study by \cite{Pinto2016} found that learning over a number of tasks helps discover richer representations of the environment, thus outperforming models that have otherwise been trained on a single task alone. But an open research problem is to consider methods by which robots can motivate themselves to select useful tasks to learn from. 
	
	In a study with ideas analogous to adversarial training, \cite{Pinto2016b} demonstrated that having a protagonist-antagonist paradigm resulted in more effective learning of visuomotor policies. They discovered that having an antagonist robot that aimed to prevent the protagonist from grasping, resulted in learning higher performance grasping, due to the necessity to learn a robust policy to overcome this adversary. In summary, they emphasize that not all data is the same. Contrary to the massive $800,000$ grasping dataset presented by \cite{Levine2016}, they found systems that attack harder examples tend to achieve faster convergence and higher performance. 
	
	\subsubsection*{On Abstract Parametrized Skills --} End-to-end methods produce amazing results by learning visuomotor torque-level control policies straight from raw pixel information. However, two immediate drawbacks are critical for autonomous systems. Firstly, the robot can only learn very task specific motions rather than abstract notions of skills and representations reuseable throughout its lifetime. In particular, GPS allows the robot to quickly encode visuomotor policies by guided trajectories acquired through optimization, but since they operate over joint torques it is difficult to decipher abstract skill boundaries\footnote{In fact, a coalition of researchers during a Robotics: Science and Systems (RSS) workshop entitled \emph{Are the Sceptics Right? Limits and Potentials of Deep Learning in Robotics} (June 2016) in Michigan, USA have argued that it may not be ideal to learn end-to-end. They indicate that in the same way that we would never want to learn sort when we have quicksort, it makes little sense to learn low-level torque activations when we understand kinematics.}. Secondly, by operating over joint torques, the network loses control guarantees and parametrization insight that abstract skills derived from control-theoretic approaches may provide\footnote{It is entirely untrue to state that Guided Policy Search algorithms have \emph{no} guarantees. Actually, its original formulation has asymptotic local convergence guarantees. Later \cite{Montgomery2016} reformulated GPS under approximate mirror descent to find convergence guarantees in simplified convex and linear settings and bounded guarantees in the nonlinear setting.}. 
	
	By learning at the lowest level of motor units (e.g. in configuration space over joint torques), systems need a massive set of examples and training steps to cover this space. For instance, even for a simple 2D spatial reaching tasks in a confined $40cm\times30cm$ area using a $6$ DOF Kinova arm with three fingers was said to require over $50$ million steps via an end-to-end (raw pixel to joint space mapping) paradigm \cite{Rusu2016b}. For such reasons, learning over a parameterized space abstracts away these basis motor units and we theorize will accelerate learning. 
	
	The notion of control guarantees is of chief importance through an industrial and product-delivering perspective. Especially in scenarios where human factors are involved or other high fidelity situations, systems that acquired expertise through learning over a history of experience must exhibit certified guarantees. Indeed, the maximum activation visualization approach, ``deep dream'' can be used to identify convolution features to attempt to make sense of the network's latent representations \cite{Mahendran2016}, but, we are concerned with a stronger sense of guarantee, especially, in the sense of control derived from the output of these networks. In autonomous driving, for example, the system must be analyzed that one can establish certified guarantees, up to sensor noise, that regardless of input, the vehicles will never attempt to give commands that result in collision with entities in the world. Such analytics is extremely difficult to reason over if the commands are over low-level specifics like wheel torques. Rather, analysts can reason easily in Cartesian space---perhaps then, the robot should learn sets of abstract skills that operate with goals that are easily interpretable for validation and certifications. As such, the direction in which we should look into may live closer to the realm of acquiring such parametrized skills.  
	
	While there exists a plethora of work for learning these skills, (e.g. \cite{Konidaris2009,DaSilva2012,Masson2015}), we believe that to better investigate a principled formulation for lifelong learning cognitive systems, we should investigate the perspective of learning through the lens of cognitive psychologists---such allows us to better understand the development of cognition and action in living organisms. Insight from this becomes fundamental in drawing computational analogs originating from developmental processes to better design artificial, learning systems. 
	
	\section{Complex Sensorimotor Hierarchies}
	The artificial neural networks architecture as an explanatory means to a connectionist model of cognition and action is not a concept that resides solely in computation. In fact, a number of cognitive psychologists showed intrigue when these networks were at its infancy, dating back two decades \cite{Rumelhart1998}. Most prominently, \cite{Thelen1996} describes that such models are exciting in the sense that there exists only process. They indicate that the essence of behavior is distributed among numerous individual units, that together, in the strengths of their connections, describe behavior. They are plastic and modify themselves through dynamical processes with the world. In particular, \cite{Thelen1996} argue against the notion that these ``neural networks contain some privileged icon of behavior, abstracted from complex motivation and environmental contexts in which it is performs.'' In other words, the theory that networks encode a particular context-independent behavior---something like a Central Pattern Generator (CPG) is entirely incorrect. They emphasize that behavior is context-specific, even in the case of CPGs---many studies that elicit such behavior and draw such conclusions are based off an impoverished form of induced behavior. To this regard, it may be true that behavior exists in some innate form that when given appropriate stimuli will generate seemingly high level actions---resulting in the misclassification of there being such generators. Thelen and Smith conclude that the development of these into complex behavior is entangled in motivation and context-specificity. 
	
	Following this notion, the work that has been discussed to this point fail to tackle this intertwined cobweb of action, environment, and motivation. Although reinforcement learning paradigms do describe the acquisition of action through the system's interaction with entities in the world in context-specific situations, many of these studies fail to indicate principled motivators. Rewards are generally task-specific and user defined---\emph{not} an inherent property derived by the system itself in its interpretation of \emph{situational contexts}. Admittedly, even promising developmental frameworks like the ones outlined by \cite{Mugan2012,Grupen2005}, are culprit to ad hoc reward structures. Most importantly, however, through \cite{Grupen2005}'s outline of \emph{figurative schematas} that organize into the development of robot behavior, they emphasize the need for control knowledge to be represented in a manner that supports generalization. Similar aspirations are found in the \emph{action schema} framework \cite{Platt-ICDL06}. The ability to construct and reused learned behaviors in a general manner is of fundamental importance---thus, we share a similar view on the acquisition of motor behavior. These views were originally derived from the proposal under \cite{Piaget1952}'s account, where human infants exhibit a sensorimotor stage that lasts approximately $24$ months while producing control knowledge that support generalization and reuse. Such reuse and organization implies underlying hierarchies of motor behavior. 
	
	In comparable work, \cite{Heess2016} emphasized the importance of hierarchical controllers that operate at different time scales in support of modularity and generalizability. Their work showed a promising step forward in locomotive skill transfer between a number of simulated bodies with many degrees of freedom, wherein high-level controllers modulated low-level motor skills which emerged from pretraining. Other work has looked into building \emph{implicit plans} or macro-actions by interaction alone using a recurrent neural network structure \cite{Mnih2016}. However, these works do not necessarily discuss how these temporal hierarchies of options, skills or macro-actions play with intrinsic motivators, where systems derive their own reward paradigms and build continuously extended motor hierarchies. \cite{Kulkarni2016} presented a hierarchical-DQN framework which integrated hierarchical value functions and intrinsic motivation by having a top-level function learn policies over intrinsic goals and a lower-level learning policies to achieve these goals. They suggested that intrinsic motivation be derived from the space of entities and relations which is sufficiently bounded and finite in Atari games, however, may exhibit explosive growth in the physical world. Future work indicated a connection to deep generative models---wherein, in this paper, we derive intrinsic motivation from a deep generative dynamics model.
	
\begin{figure}
		\centering
		\includegraphics[width=0.485\textwidth]{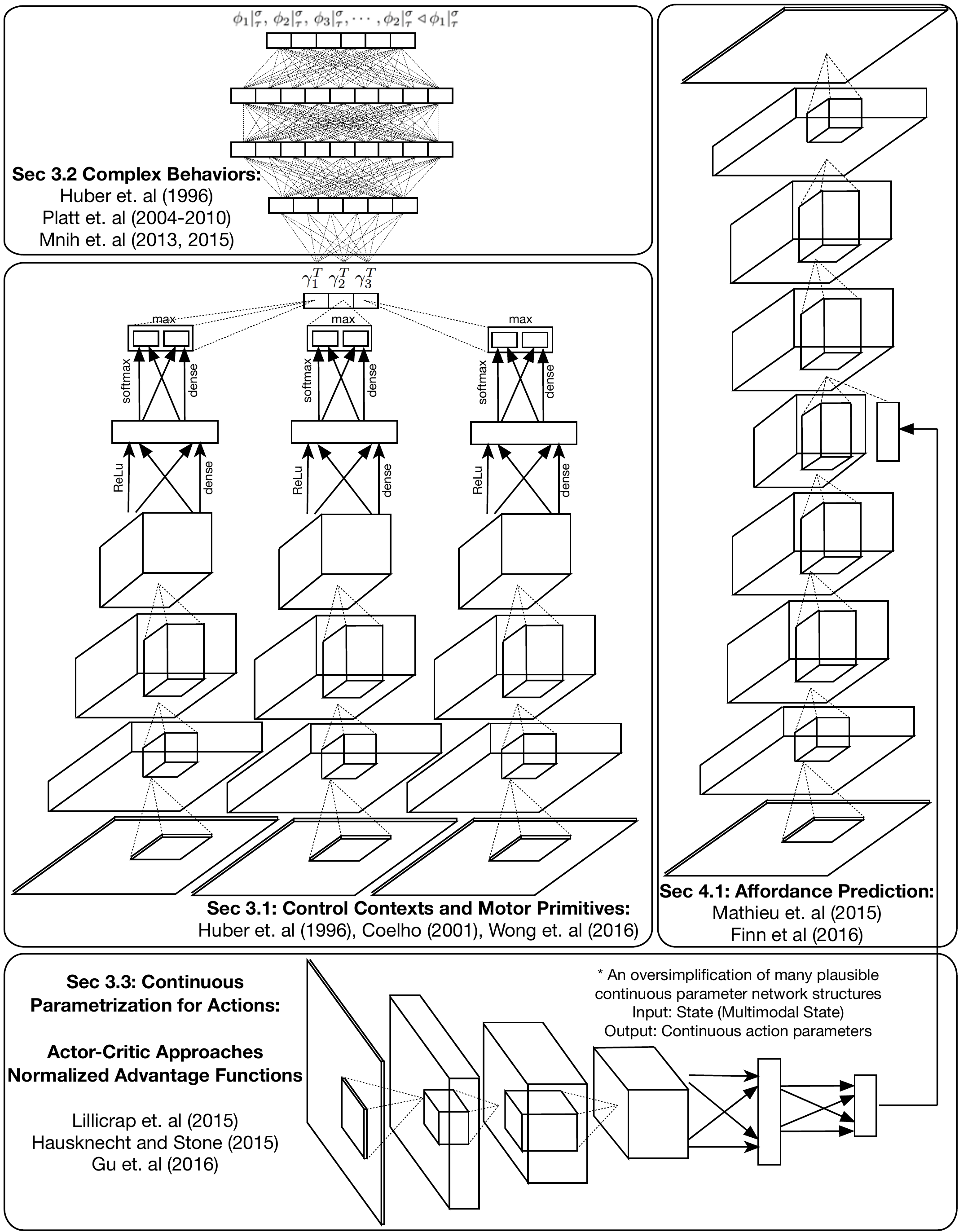}
		\vspace{-17.5pt}
		\caption{A candidate framework marrying various learning modules that have been individually presented in the literature}
		\label{fig:overall}
	\end{figure}

	In the following subsections, we summarize work that provide systems with the ability to learn complex sensorimotor hierarchies resulting from experience and interaction with the world. Complex action-related behavior are expressed as motor hierarchies emerging through the combinatoric sequencing and composition of actions, that at the lowest level, are learned by associating sensory input to resolve motor primitives. As such, we begin at the lowest level of motor development, on how a robot can associate sensor input to activate closed-loop motor primitives. Next, we investigate learning to bootstrap these primitives to develop more complex behaviors. And lastly, we incorporate techniques present in the literature to suggest the learning of control goals that evolve primitive reflexes to intentional goal-oriented behaviors. In Section~\ref{sec:affordance}, we describe an intrinsically motivating paradigms that leverages the system's ability in its understanding of the world and of its own inherent actions and representations through \emph{control contexts}---such motivators are to drive the processes that govern the development of these sensorimotor behaviors and cognitive representations. A plausible unifying framework is illustrated in Figure~\ref{fig:overall} by piecing together various selected studies currently in the literature. 
	
	We now quickly elaborate on each of these pieces and on their respective subsections in this manuscript. 
	
	Section~\ref{subsec:prims} describes how to activate motor primitives given sensory input, deriving a \emph{control context}---it discusses an approximation to the function $f_{\gamma_i^T}: s\mapsto \gamma_{\phi_i|^\sigma_\tau}^T$. 
	
	Section~\ref{subsec:complex} investigates how to build hierarchies of complex behaviors by combining existing controllers to construct new ones under the current state description $\Gamma^T(s) = \gamma_1^T \cup \gamma_2^T \cup \cdots \cup \gamma_n^T $. This is described by learning the policy $\pi_{n+1}^*: \Gamma^T(s)  \mapsto \{\phi_1|^\sigma_\tau, \phi_2|^\sigma_\tau, \cdots, \phi_n|^\sigma_\tau\}$. The networks proposed here approximates the value function $f_{\pi_{n+1}^*}: \Gamma^T(s) \mapsto V^{\pi^*_{n+1}}(\Gamma^T(s))$, where $\Gamma^T$ is the deep control context at time $T$ encompassing all control state descriptions $\gamma^T$. 
	
	Section~\ref{subsec:param} describes methods to learn continuous control parameters for controllers $\phi_1|^\sigma_\tau, \phi_2|^\sigma_\tau, \cdots, \phi_n|^\sigma_\tau$, thus approximating the function $f_{\rho_i}: s \mapsto \rho_{\phi_i|^\sigma_\tau}$.
	
	Since both the learning of $f_{\pi_{n+1}^*}$ and $f_{\rho,\phi_i|^\sigma_\tau}$ is by trial and error, it requires that environmental reward is defined. We refer to the system's level of understanding regarding entities in the world and of its own actions to establish  reward. As such, Section~\ref{subsec:dynamics} discusses learning to predict the transition dynamics of the world by approximating $f_{W}: s, \rho_{\phi_i|^\sigma_\tau} \mapsto s'$. 
	
	Section~\ref{subsec:reward} outlines a plausible reward structure derived from affordance predictions. 
	
	And finally, Section~\ref{sec:planning} describes how a system can exploit these learned behaviors and representations to solve useful tasks and mission in the world.

	\subsection{Activating Motion Primitives}
	\label{subsec:prims}
	In their revolutionary book, \cite{Thelen1996} outlines the misinformations of contemporary theories in cognitive development, suggesting that it is not the case that cognitive and motor skills emerge linearly through development but these primitive forms of behaviors are inherently ingrained. They are reinforced by interaction with the environment to seek dynamically stable solutions. As a result, skills emerge from context-specific situations that are afforded by the world. It appears that local circuitry within the spinal cord mediates a number of closed-loop sensory motor reflexes---for instance, the spinal stretch reflex \cite{Purves2001}. In fact, it has been observed that all humans developing infants exhibit similar chronicles of reflexive motor behaviors. 
	
	The Central Nervous System is organized according to \emph{movement patterns} \cite{Aronson81}---with its most basic form being the reflex\footnote{\cite{Grupen2005} expresses that \emph{packaged movement patterns} ``reside in the central and peripheral nervous system and range from involuntary responses to cortically mediated visual reflexes [and] contribute to the organization of behavior at the most basic level by constituting a sensorimotor instruction set for the developing organism.''}. Under the notion of epigenetic developmental theory primitive reflexes, expressed as neuro-anatomical structures, are the basic building blocks of behavior. In this regard, complex behavior emerges from combinatoric sequences of primitive control in response to reinforcement by the world. These primitive forms of behavior collectively describe an epigenetic computational basis by which complex control actions are derived \cite{Grupen2005}. The ability to convert these developmental reflexes into intentional actions remains an fundamental cognitive process \cite{Zelazo1983} and has been briefly explored in work by \cite{Wong-RSS2016} in which a deep network controlled closed-loop motion primitives. This was accomplished through the activation of output neurons describing control state derived from time varying dynamics. In short, this reduces sensorimotor control to a supervised learning problem in which the state of the world described through sensor modalities (vision was shown in their study) is used to predict the probability of controller state transitions. Consequently, additional complexity of output neurons simplified learning to a reduced network size and training data.  
	
	The primitive controllers used in their study are all instances of negative feedback stabilized systems. In other words, these motor primitives are regulators that minimize error between sensation and desired. Consider a simple proportional derivative heading ($\theta$) controller that yields a single degree of freedom dynamical system represented by the second order differential equation in the canonical form, 
	\[\ddot{\theta} + 2\zeta\omega_n\dot{\theta} + \omega_n^2\theta = 0\]
	where $\zeta = B/(2\sqrt{KI})$ is the damping ratio and $\omega_n=\sqrt{KI}$ is the natural frequency. It is assumed here that $K, B, $ and $I$ are all constants representing the proportional and derivative gains, and the scalar moment of inertia around the rotation axis respectively under a canonical spring-mass-damper model \cite{Hebert2015}. Closed-loop feedback controllers such as these have well understood proofs of stability. For instance, the proportional derivative (PD) controller such as the one we described here is provably stable \cite{Lyapunov1992}. This property is established for systems formulated as harmonic oscillators similar to the above second order differential equation. Convergence results have been proven for closed-loop controllers by \cite{Coelho_JRS97} for regular convex prismatic objects when two closed-loop controllers executed in a particular sequence. Experiments were later shown on a robot manipulator agreeing with such convergence guarantees \cite{Coelho-Thesis01}\footnote{In the next subsection, we will discuss composite controllers which have similar convergence guarantees as outlined and proven by \cite{Platt-TRO10}.}. Likewise, optimal controllers described by regulators like linear quadratic regulators (LQRs) and its variants fall into this categorization as well. In fact, dynamically balancing robots like the uBot platforms \cite{Kuindersma09:KnuckleWalking,Ruiken2013}, implement a variant of LQR. To reiterate the notion of learning abstractions, we suggest strongly that rather than learning a balancing policy from scratch, perhaps a better direction is to consider this closed-loop controller as a skill and employing learning architectures at the skill level of abstraction. 
	
	Namely, in spirit of the flurry of work on autonomous vehicles, we emphasize that motion planners and path tracking procedures generally fall under this umbrella notion as well. In particular, path tracking in navigation at the most primitive sense, leverage a heading and longitudinal controller like those described by \cite{Hebert2015}. Potential field methods for path planning \cite{Ge2002,Wang2000,Khatib85} like Harmonic function path planning \cite{Connolly1993} have shown success dating back over two decades of research. Especially with recent advances in GPU parallelization for efficient relaxation and the logarithmic transformations to prevent diminishing gradients \cite{Wray2016}, these classical methods still remain powerful path generators. An elegant control theoretic relaxation-based method to velocity planning presented by \cite{Hebert2015} is shown to be done in linear time of the path, resulting in minimal path deviations and maximal performance envelope. Many of these navigation solutions are fast and robust---so instead of replacing them completely with a learned approximation, we suggest treating resulting plans like motion primitives or inherent behaviors in the context of forming complex visuomotor hierarchies. Likewise, solutions to other motion planners like RRTs or A* variants can interpreted as motion primitives given that their trajectory can be used to describe some transient, converged, or goal completion evaluation. 
	
	Under this broad encompass, even dynamic motion primitives (DMPs) \cite{Ijspeert_NIPS03} and other powerful optimization based methods for locomotion, like those developed during the DARPA Robotics Challenge (e.g. \cite{Feng2015b,Kuindersma2016}), fall into this categorization of plausible motor primitives.
	
	Motion primitives like these can be formalized under the \emph{Control Basis framework} in which the interaction between the embodied system and the environment is modeled as a dynamical system, allowing the robot to evaluate the status of its actions as a state describing a time varying control system. These controllers  $\phi|^\sigma_{\tau}$, consist of a combination of potential functions ($\phi \in \Phi$), sensory inputs ($\sigma \subseteq \Sigma$), and motor resources ($\tau \subseteq \mathcal{T}$) \cite{Huber1996}. Controllers achieve their objective by descending along gradients in the potential function $\nabla\phi(\sigma)$ with respect to changes in the value of the motor variables $\partial{\sbf{u_\tau}}$, described by the error Jacobian $J=\partial\phi(\sigma)/\partial{\sbf{u_\tau}}$. References to low-level motor units are computed as $\Delta u_\tau = \kappa J^\# \Delta\phi(\sigma)$, where $\kappa$ is a control gain, $J^\#$ is the pseudoinverse of $J$ \cite{Nakamura1990}, $\Delta\phi(\sigma)$ describes the difference between the reference and actual potential \cite{Sen2014}. 
	
	The time history or trajectory of dynamics ($\phi, \dot{\phi}$) as a result of interactions with the environment by executing controllers have been shown to have predictive capability regarding the state of the environment. It was originally shown by \cite{Coelho1998} that dynamics elicited by abstract actions in the form of controllers serve as important identifiers for the current \emph{control context}---one of many finite sets of dynamic models that capture system behavior. The state description $\gamma^t$  for a particular control action $\phi|_\tau^\sigma$ at time $t$ is derived directly from the dynamics ($\phi, \dot{\phi}$) of the controller given a specified control goal $g$ such that, 
	\begin{align*}
	\gamma^t(\phi|_\tau^\sigma) = 
	\begin{aligned}
	\begin{cases}
	\mbox{\textsc{Undefined}}: &\phi \mbox{ has undefined reference} \\
	\mbox{\textsc{Transient}}: &|\dot{\phi}| > \epsilon\\
	\mbox{\textsc{Converged}}: &|\dot{\phi}| \le \epsilon\mbox{ and } \phi \mbox{ reaches } g  \\
	\mbox{\textsc{Quiescent}}: &|\dot{\phi}| \le \epsilon\mbox{ and } \phi \mbox{ fails to reach  } g
	\end{cases}
	\end{aligned}
	\end{align*}
	Collectively, the state descriptions over all abstract actions describe the control context. This collection can be thought of as a compressed variant of sensorimotor contexts \cite{Hemion2016b} that is specifically grounded to particular control interactions through the function approximate $f_{\gamma^T}:s \mapsto \gamma^T_{\phi|^\sigma_\tau}$, however, the hierarchy in this case, is achieved by learning new control programs to discover more abstract contexts. 
	
\begin{figure}
		\centering
		\includegraphics[width=0.475\textwidth]{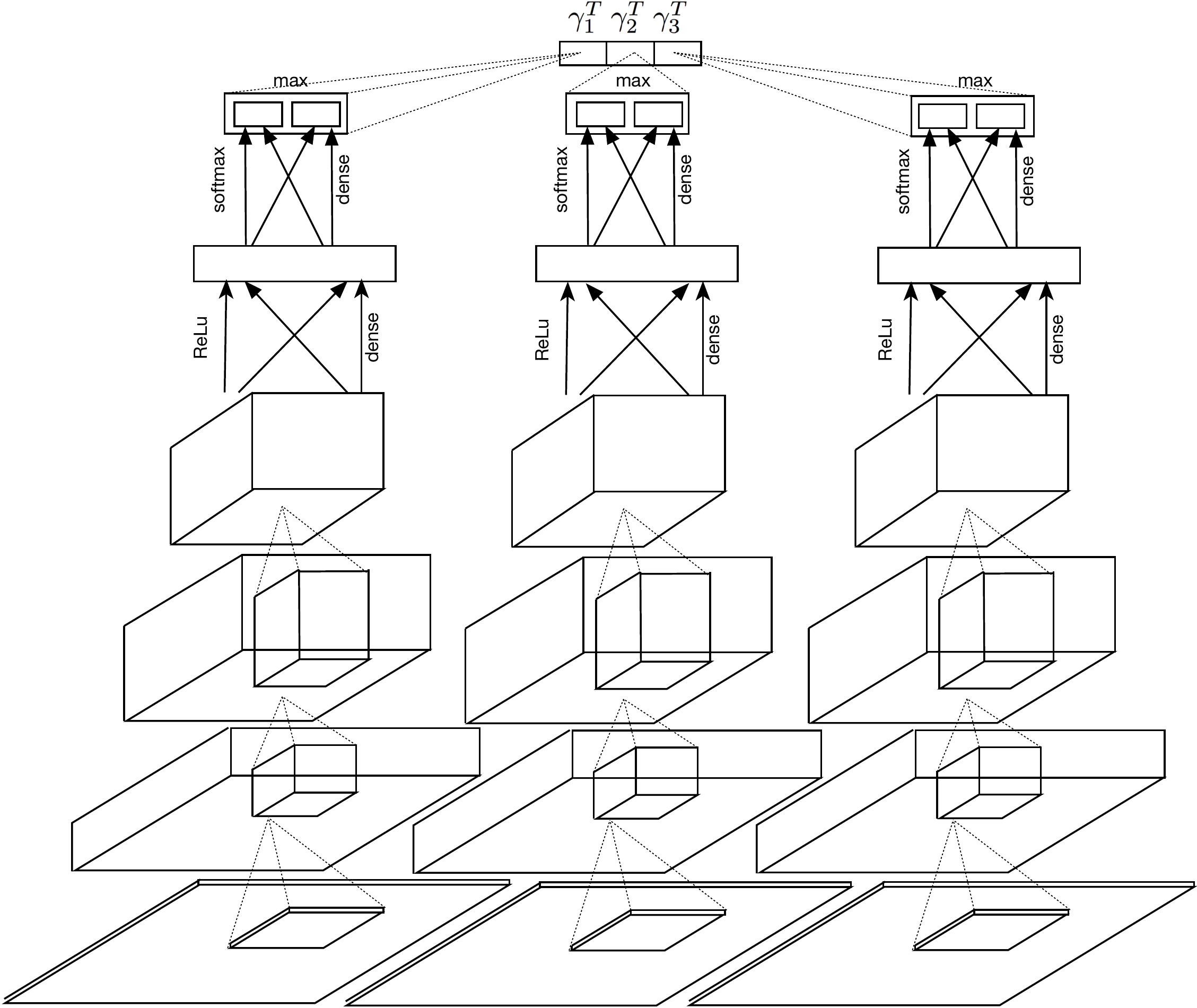}
		\caption{A \emph{deep control context} defined over the collection of control state predictions from each $\gamma^T$-network}
		\label{fig:control-context}
	\end{figure}

	Motor units that operate at the torque or joint level are abstracted away into high level parameterized (continuous and goal-oriented) motion controllers that achieve particular objectives. As such, a surrogate of control context is produced by inferring the time-varying dynamics---specifically, Figure~\ref{fig:control-context} illustrates the concatenation of state descriptions over several $\gamma^T$-networks originally presented by \cite{Wong-RSS2016} to formulate a \emph{deep control context}, which provides implications into intriguing reinforcement learning methods that have been otherwise inubiquitous.
	
	\subsection{Composition of Motor Behaviors}
	\label{subsec:complex}
	Learning these $\gamma^T$-networks results in a set of sensorimotor policies for activating motor primitives---such a policy is a sensory-driven predictive process that activates primitives when appropriate state descriptions are predicted by the network. A fundamental drawback is that this reduces the set of available behaviors of a system to some static set preordained by initial primitive ``reflexes''---in fact, it is indeed the size of the primitive set. This is not the case in development however, since the set of skills and their competence grows \cite{Thelen1996}. As such, both robots and humans must learn how to make use of their innate primitives to build new, complex motor behaviors. 
	
	Previously, we discussed how infants derive complex actions through developmental chronicles of emergence and inhibition of primitives. In the same sense that complex motor behaviors emerge from combinatoric sequences of primitive reflexes in the human infant, a computation composition and sequencing of primitive closed-loop motor units form policies for high-level control policies. 
	
	This idea is not new. The idea of acquiring hierarchies of motor policies through reinforcement learning over a basis of existing motor controllers dates back two decades to work by Huber et. al (See, \cite{Huber-ICRA96,Huber-IJCAI97,Grupen2005}). This framework was first shown on a quadruped walking robot, where it quickly learned walking gait behaviors given a set of primitive kinematic conditioning controllers in a state space constrained by valid tripod stances. They outlined the learning of a locomotive gait schemata over a basis of quasistatic kinematic conditioning controllers with time-varying constraints represented as artificial maturation processes and domain requirements. Unfortunately, these constraints are defined a priori and are inherent in their developmental scheduler or assembler. Rewards are closely tied to phases of development and known task specifications. For instance, in particular developmental stage the reward attributed to the maximization of angular velocity. Regardless, the system was able to learn a compilation of control knowledge through environmental, task relevant rewards under traditional $Q$-learning over the control context or concatenated state descriptions over the basis of innate control programs. Emerging schemata consisted of rotate, translate, and maze traversal behaviors. Expressiveness was increased by using a method to allow for concurrent control composition \cite{Grupen2005}. Later, this was extended to a number of other applications, for instance in robot manipulation and grasping domains producing a series of profound and sophisticated control paradigms namely in work by \cite{Platt_IROS02,Platt_ICRA04,Platt-Humanoids05,Platt-ICDL06,Platt-TRO10}. To clarify, the forming of new control policies can be formulated under the \emph{options} framework \cite{Sutton-AIJ99,Stolle2002}, where primitive sensorimotor skills are seen as possible high-level abstracted options. 
	
	New control policies do not emerge by sequencing innate controllers alone, but also can emerge by \emph{combining} many controllers, or simply operating inferior control programs inside the nullspace of primal ones. An important aspect to the creation of new control policies is the notion of \emph{null space composition}, a mechanism for systematically specifying composite controllers derived from combinations of other controllers. To our knowledge, this was first introduced in work by \cite{Huber-ICRA96} and popularized in manipulation by the work of Platt. This composition uses the \emph{subject-to}, $\triangleleft$, operator to project the control gradient of subsequent, subordinate controllers into the null space of a superior controller, allowing all controllers to attempt to achieve their objectives if there exists resources allowed by those superior in a sequential priority fashion. Thus, new control policies can be expressed in terms of existing motor behaviors. \cite{Platt2006} derives a general formulation for the composition of $k$ controllers in numerical priority under the effector space $Y$ as the following, 
	\vspace{-2.5pt}
	\begin{align*}
	\begin{aligned}
	\nabla (\phi_k\,\triangleleft\, ,\hdots, \triangleleft \,\phi_1) & \equiv  \, \nabla_y\phi_1 + [\mathcal{N}(\nabla_y\phi_1^T)]\nabla_y \phi_2 \\
	&+ \Big[\mathcal{N}\Big(\genfrac{}{}{0pt}{0}{\nabla_y\phi_1^T}{\nabla_y\phi_2^T}\Big)\Big]\nabla_y\phi_3  \\ 
	& \,\,\, \vdots \\
	& + \Bigg[\mathcal{N}\Bigg(\begin{array}{c}
	\nabla_y\phi_1^T \\[-5pt]
	\vdots \\[-2pt]
	\nabla_y\phi_{k-1}^T
	\end{array}\Bigg)\Bigg]\nabla_y\phi_k
	\end{aligned}
	\vspace{-15pt}
	\end{align*}
	The null space of the $(k-1)$ controllers are computed by concatenating all control gradients as, 
	\begin{align*}
	\begin{aligned}
	\mathcal{N}\Bigg(\hspace{-5pt}\begin{array}{c}
	\nabla\phi_1^T (y)\\[-5pt]
	\vdots \\[-2pt]
	\nabla\phi_{k-1}^T(y)
	\end{array}\hspace{-5pt}\Bigg)= I - 
	\Bigg(\hspace{-5pt}\begin{array}{c}
	\nabla\phi_1^T(y) \\[-5pt]
	\vdots \\[-2pt]
	\nabla\phi_{k-1}^T(y)
	\end{array}\hspace{-5pt}\Bigg)^{\hspace{-4pt}\#}
	\Bigg(\hspace{-5pt}\begin{array}{c}
	\nabla\phi_1^T(y) \\[-5pt]
	\vdots \\[-2pt]
	\nabla\phi_{k-1}^T(y) \end{array}\vspace{-10pt}\Bigg)
	\end{aligned}
	\end{align*}
	where $I$ is identity. The above produces a null-space $\mathcal{N}$ that is tangent to all $k-1$ gradients. The idea of operating multiple control programs through compositions in null-space paves way for coarticulate control paradigms. This notion relies on the fact that in many cases controllers are redundant, having more resources than necessary to achieve a particular objective (e.g. redundant in the sense of degrees of freedom). As such, these excess resources can be used by inferior controllers to achieve secondary objectives simultaneously by operating under the constraint of not interfering with the primary objectives \cite{Rohanimanesh2004}.                                             
	
	The optimal sequencing of controllers described by some policy $\pi$ to solve a particular task is generally achieved through table-based reinforcement learning using the control context as a state representation. Yet, the use of neural networks for classical control problems like the peg-in hole insertion task dates back to the early nineties. For instance, work by \cite{Gullapalli1992} used multilayered feedforward networks to generate real valued outputs as control parameters. Simple networks such as these can easily approximate the value functions that describe the optimal policy $\pi$ that expresses a trajectory through control contexts and closed-loop (primitive and composite) controllers. In fact, this is a simplified version of DeepMind's Atari approach \cite{Mnih2013,Mnih2015} where current state is used to approximate a value function, however the key distinction is that by describing current state via control contexts, this reduces the input dimensionality drastically. The options framework has been demonstrated with DQNs where option heads were integrated into the output end of the networks. \cite{Osband2016} used these ``bootstrap heads'' with different motivations allowing the network to output a distribution over Q-functions. Similarly, this was later extended by \cite{Arulkumaran2016} with an additional supervisory network, allowing the policy to be decomposed into combination of simpler sub-policies. Instead, since we learn over control contexts the input dimensionality is extremely reduced and leads to faster learning. 
	
	If one were to consider the use of these $\gamma^T$-networks as a perceptual interface to activate motor primitives, null-space composition is achieved simply by taking the controller $\phi_i|^\sigma_\tau$ corresponding to the control state $\gamma_i^T$ predicted by the network and using the null-space operator between multiple instances of these (e.g. the composition of networks $f_{\gamma,\phi_j|^\sigma_\tau}$ subject to $f_{\gamma,\phi_i|^\sigma_\tau}$ is trivially $\phi_j|^\sigma_\tau \triangleleft \phi_i|^\sigma_\tau$). 
	
	Each controller $\phi|^\sigma_\tau$ requires a control reference that implies an error to minimize. In many cases, these goals are generally human-defined. For instance, \cite{Hart2009thesis} uses hue saturation to select interesting areas in the scene. However, recently deep learning architectures were shown to be powerful tools of extracting candidate features in the world. Leveraging the predictive $\gamma^T$-networks provides these indicates stimuli in the world that derives useful control goals, or in other words, control references, that feed into these controllers $\phi|^\sigma_\tau$. In fact, the null-space composition in this case, is mostly unchanged and operate almost identically. A forward propagation over each $\gamma^T$-network predicts control context or state description that indicate activations for each primitive. 
	
	In the following section, we investigate the literature in reinforcement learning, both deep and classical, for methods that learn these control references or goal parameters that take on continuous values. 
	
	\subsection{Continuous Control Parameterization}
	\label{subsec:param}
	
	The $\gamma^T$-network shown in \cite{Wong-RSS2016} makes an unfortunate assumption that the motion primitive described by closed-loop controller $\phi|^\sigma_\tau$ inherently defines a static control goal $g$. This particular form of goal is derived from cognitive development insight, allowing networks to learn \emph{when} to execute particular abstract skills represented as reflexive actions. However, the question that remains is the encoding of \emph{how} should these skills be performed. Infants readily bootstrap their innate reflexive repertoire to quickly learn the functionality of their end effectors as entities conform to their hands via palmar grasp reflex.  Static parametrized goals allows for quick association-based learning, but do not generalized well to future task that require goals outside of innate reflex descriptions. As a result, both infants and robots must learn useful parameters to their inherent motor behaviors in response to stimuli in the world. The $\gamma^T$-networks can be extended to predict varying parameterizations of control goals by a concatenation of state and action parameters after the convolutional layers, a technique that is used in many predictive networks like those presented in \cite{Levine2015,Finn2016b}. In fact, work by \cite{Takahashi2017} implemented this extension producing a variant of $\gamma^T$-networks that account for varying control goal parameters. 
	
	Learning control parameters is generally accomplished via a trial and error approach, for instance, through popularized deep reinforcement learning methods \cite{Mnih2013,Mnih2015}. Unfortunately, a major drawback of the approach used to tackle learning control policies for the Atari games is that the network's output is fixed over the set of discrete control actions. Quite evidently, robots must be able to perform continuous parameterized actions in the real world instead of fixed discretized control actions. To address this \cite{Lillicrap2015} presented an actor-critic approach to learn over continuous domains with neural networks using a deep variant to the deterministic policy gradient originally proposed by \cite{Silver2014}. Similarly, \cite{Balduzzi2015} also extended the original deterministic policy gradient algorithms with a network that explicitly learned $\partial Q/\partial a$---unfortunately though, their methods were only shown on low-dimensional domains. 
	
	\cite{Heess2015b} introduced the stochastic value gradients to learn stochastic policies through a $Q$-critic. They found that stochastic control can be supported by treating the stochasticity in the Bellman equation as a deterministic function consisting of external, Gaussian noise. From this, they revealed a ``reparametrization'' trick, similar to that of \cite{Kingma2013}. Meanwhile, \cite{Wawrzynski2013} trained stochastic policies using a replay experience buffer with the actor-critic framework. The use of the replay buffer has been essential to ensure that data samples are independent and identically distributed. This particular technique was popularized by \cite{Mnih2015} with original insight dating back twenty years in work by \cite{Lin1993}. A trusted policy optimization approach proposed by \cite{Schulman2015}, directly builds a stochastic neural network policy without this decomposition and does not require the learning of an action-value network. It appears to produce near monotonically improvements but require careful selection of updates to the policy parameters to prevent large divergences to the existing policy. Furthermore, it has been theorized that this technique appears to be less data efficient. Work by \cite{Hausknecht2015} has demonstrated a solution to reinforcement learning in continuous parameterized action spaces, where they successfully trained a RoboCup soccer agent that scored more reliably than the 2012 champion. 
	
	Using value function estimation like these approaches for continuous domains generally uses two networks to represent the policy and value function individually \cite{Schulman2015,Lillicrap2015}. There has been work however, to reformulate the original $Q$-learning scheme that results in an elegant effort that can be ported to the continuous setting---namely work by \cite{Gu2016} has attempted to show this by learning a single network that outputs both policy and value function. Their work was based off dueling networks shown by \cite{Wang2015} where they decomposed learning into two streams corresponding to the learning of state-value and advantages for each action---though their work was presented in the discrete setting. These \emph{dueling networks} were improved from the formulation of \emph{double $Q$ networks}. The motivation behind double $Q$ networks is that $Q$-learning, even in the tabular setting, exhibits overoptimism due to estimation errors, however, \cite{vanHasselt2015} found that by decomposing the max operation in the target into two value functions corresponding to action selection and action evaluation, not only reduced this particular overoptimism, but also lead to performance increase. 
	
	However, approaches that rely on a model-free reinforcement learning method like traditional $Q$-learning has yet another major drawback. They are tragically inefficient with experience. For example, the learned Atari policies required millions of gameplay examples to converge to competence. Consequently, in the past a number of model-based methods like Dyna \cite{Sutton1990,Sutton1991}, Prioritized Sweeping \cite{Moore1993}, and Queue-Dyna \cite{Peng1993} have been suggested to make more efficient use of training examples while increasing computation. Many of these select $k$ samples to use for update as opposed to a single update with traditional $Q$-learning. To do this, these methods use experience to not only learn an optimal policy, but also construct a transition $\hat{T}$ and reward $\hat{R}$ model, meanwhile updating the values of $k$ additional state-action pairs. And as summarized by \cite{Kaelbling1996}, Dyna does this by selecting $k$ randomly while the latter two methods prioritize the selection of the pairs to ``regions of interest.'' Dyna is shown to converge ten times faster than traditional $Q$-learning and the prioritized methods being two-fold faster then Dyna---thus, making much better use of experiences for learning. Quite recently, these ideas stemming back two decades of research has reincarnated into a deep learning framework that incorporates model-based acceleration for deep reinforcement learning with continuous control parameters. First, \cite{Gu2016} reformulated $Q$-learning in the continuous setting into \emph{normalized advantage functions}, an alternative to policy gradient and actor-critic methods, which decomposes the quality term $Q$ into a state value term $V$ and an advantage term $A$. This particular insight have been explored by others in the past \cite{Baird1993,Harmon1996,Wang2015}. Next, they showed that policy learning can be achieved by taking a learned model of the dynamics and simulating synthetic plausible outcomes via \emph{imagination rollout} and appending the experiences to the replay buffer. Doing so, increases the efficiency of data usage and is a likely candidate to learn the control parameters needed for $\gamma^T$ networks. Interestingly, these imagination rollouts according to learned dynamics models corresponds to a form of $\lambda$-return, where given this model, we can simulate a number of $n$ step trajectories by traversing this aspect transition network. We then weigh these $n$ step backups yielding a compound backup---such an update has been shown to make more efficient use of experiences \cite{SuttonBarto98}. 
	
	In a distributed approach, \cite{Gu2016b} introduced a parallelizable learning algorithm leveraging the normalized advantage functions, to be used across multiple robots which can pool their policy updates asynchronously resulting in accelerated learning. 
	
	With almost all frameworks to date, experience updates from the replay buffer were uniformly selected from this replay buffer. In fact, it this may not be ideal since individual samples may have varying degrees of significance. As a result, \cite{Schaul2015} proposed the \emph{prioritized experience replay} which samples  at the same frequency they were originally experienced---this result was shown to improve state of the art, and perhaps could be a potential candidate for model-based acceleration sampling. Similar works by \cite{Zhai2016} used prioritized sampling to bias experience selections.   
	
	Model referred updates accelerates learning by simulating synthetic on-policy futures. In essence, this describes a forward dynamics model of the interactions, perhaps even over a lifetime of interaction experiences. Evidently, the use of Guided Policy Search tries to fit the dynamics in a locally-linear fashion and using the model as a reference to apply imagination rollouts by placing these cumulatively constructed synthetic experiences into the replay buffer for updates. There is in fact, a connection between these dynamics models that attribute to the increased efficiency of deep reinforcement learning methods and algorithms to provide artificial curiosity. Quite fortunately, these representations that concern the transition dynamics and reward have immediate ties and are analogous to an interaction-based knowledge repertoire that adheres to a series of task planning and intrinsic motivation frameworks. 
	
	\section{Unsupervised Affordances}
	\label{sec:affordance}
	The development of an interaction-based knowledge repertoire is an important structure that can be incorporated into a number of algorithms that necessitates forward models or predictions of future state. Fortunately, the learning of this action centric knowledge is a close analog of the transition dynamics that is \emph{already} a component of these model-based acceleration techniques presented to make efficient use of experiences in reinforcement learning paradigms. As such, this structural acquisition has immediate ties to intrinsic motivators that seek to provide artificial curiosity to autonomous systems. Intrinsic motivators allows systems to build their \emph{own} representations that reflect the inherent uncertainties of the system. 
	
	Curiosity is important for learning systems---as without a sense of curiosity, learning becomes very task specific and one dimensional---the robot can not choose to learn novel skills, only a task-specific motion for a human-defined task. For such reasons, one might consider introducing some form of curiousity \cite{Frank2015} into the approaches previously discussed. We believe that the development of motor behavior should be driven by intrinsic motivators in order to learn task-generalizable skills that can be reused in the future. While a thorough survey of intrinsic motivators is outside the scope of this paper (for further readings, see \cite{Barto2013}), we acknowledge that a number of researchers who have applied schemes to autonomously learn new skills \cite{Utgoff2002,Barto2004} and representations \cite{Oudeyer2007,Hart2009}. Instead, we envision a system that accomplishes this \emph{simultaneously}.
	
	As for this, we look into insight from cognitive development. In fact, a number of studies have shown that motor development in biological systems greatly influences the development of perception and cognition. For instance,  \cite{Piaget1953,Piaget1954}  described motor skills as a mechanism that drives development in other domains by generating new sensorimotor experiences and further studies have described cognition and perception as embodied phenomena---grounded to the body and its actions \cite{Gibson1988}. Other studies have shown that infants are highly sensitive to action-outcome relations and are capable of learning contingencies between their own behavior and outcomes in the world. In particular, \cite{Libertus2010} showed that sensory-motor experiences motivates infants to reproduce manipulation outcomes and foster reaching and grasping skills. Evidently, infants learn models through manipulation and interaction forming a sense of behavioral organization\footnote{The developmental process from neonate to approximately a year in age consists of a precisely-timed chronicle of emergence and inhabitation of primitive, postural, or bridge reflexes that contribute to the organized development of complex behavior and skill acquisition \cite{Law2011}. With age, myelination occurs in the infant, resulting in increased controllable degrees of freedom and resolution in motor activity  \cite{Oudeyer2013}. This form of maturation is especially prominent when fine motor control start to emerge in cases such as pincer grasp reflexes. When the infant develops appropriate skills, it begins to play and interact with the environment and objects in it by exploratory activity \cite{Oudeyer2007}. Increased motor acuity and refined motor skills are important for development in general and affect what kinds of information can be extracted from the environment. As motor skills develop, complicated representations of the world can too be constructed through addition information provided through these actions \cite{Libertus2010}.}. Such organization is an interplay between representation and motor development, driven by curiosity---which we regard, should adhere to model-referenced objectives that aim to explain the complex dynamic phenomena in the world. A common, perhaps, ubiquitous representation describes action-related contexts regarding entities in the world. Such a representation is of chief importance to cognitive systems. In fact, \cite{Hemion2016} argues that instead of having contemporary computational reinforcement learning agents  learn each individual skill entirely from scratch, the development of a world model can be used to support adaptive behavior and learning for cognitive systems.

	\subsection{Predicting the Dynamics of the World}
	\label{subsec:dynamics}
	Originating from ecological approaches to visual perception, the central concept to the Gibsonian perceptual framework is the notion of an \emph{affordance}, an observable environmental context that invokes a variety of latent interactions. Affordances emphasize an agent-world relationship and constitutes an interactionist account of perception as it reflects environmental signals in relation to an agent's ability to act on those signals \cite{Chemero03}.  In the strongest sense, Gibson's theory of \emph{direct perception} holds that the transformation from signal to behavior is expressed directly by neural projections that evolve to recognize opportunities for context-specific actions \cite{Reed96,Turvey92}. Such theories emphasize that percepts themselves provide a direct index into all the ``action possibilities latent in the environment'' \cite{GibsonJJ77}, thus, applicable actions and related outcomes are immediately recognized without necessarily identifying the object itself. 
	
	The Gibsonian notion of affordances describes action possibilities and can be seen as a surrogate of world state, induced by sensory input and interaction. Gibson's theory of affordance advocates for modeling the environment directly in terms of the actions it affords. These representations are idiosyncratic and reflect only those actions that can be generated by the agent. Research has been done to investigate the autonomous acquisition of such affordance representations with intrinsic motivators. For instance, an example of multiple intrinsic reward functions have been proposed to learn the transition dynamics of a particular task \cite{Hester2015}. Others have looked into domain-independent intrinsic rewards, like novelty or certainty, for learning adaptive, non-stationary policies based on data gathered from experience \cite{Hart2009,Sequeira2014}. In particular, \emph{model exploration programs} have been presented by \cite{Hart2009}, but the methods reported lacked multimodal sensor integration and do not produce knowledge structures that are easily transferrable to other tasks. A \emph{multimodal structure learning} paradigm was proposed by \cite{Wong-ICDL2016} extending the ideas initially presented by Hart---in their studies, they leveraged a promising representation describing affordances in terms of \emph{aspect nodes}. The graphical structure called an \emph{aspect transition graph} encodes Markovian state as nodes and actions as edges in a multi-graph \cite{Ku-ECCV14}. Such a model is generally used in object identification tasks by planning in belief space (rolling out a population of these forward models) to select informative actions \cite{Sen2014,Ruiken2016}. Methods for robots to autonomously acquire these models has been described in work by \cite{Wong-ICDL2016,Ruiken2016}, where systems are intrinsically motivated by a variant of the differential variance function originally proposed by \cite{Hart2009} to acquire complete graphical representations of objects---associating actions and futures derived from all possible interactions under controlled settings. 
	
	Evidently though, the aspect transition graph model has two major disadvantages. One being that in studies regarding learning these graphs by \cite{Hart2009,Wong-ICDL2016,Ruiken2016}, it is assumed that a sample mean and variance approximates the true underlying transition distribution. Unfortunately, it is not the case that this distribute is necessarily Gaussian $\mathcal{N}(\mu, \Sigma)$, for instance, it may be arbitrarily complex and multimodal. The next large criticism is that these models assume some discretization granularity of sensory input space into aspect models. The definition of what constitutes an \emph{aspect} is task-dependent and difficult to manage in a task-generic way. Both of these issues can be addressed by approximating this arbitrary complex representation in high dimensionality---this approximation over the Markovian state describing what constitutes an aspect and potential outcomes derived from interactions attributes to a new model, a \emph{deep aspect transition network}. This network is otherwise an extension of the original aspect transition model graphical structure with the key distinction being that it captures interactions over many possible granularities, deriving a continuous form of aspect state.\footnote{The choice to derive these transition networks from affordances defined through aspect graphs is due to convenience. We are aware and acknowledge the numerous affordance representations in the literature ike Object Action Complexes (OACs) \cite{Kruger2011}, etc, to name a few.} A fundamental extension to the graphical structure is that the deep variants makes no assumptions of aspect boundaries, rather, aspect nodes take on continuous state description. 
	
	The acquisition of a model that explains the dynamics of entities in the world and their evolution through interaction (the representation we referred to as an \emph{aspect transition network})  is closely related to work that attempts to predict physics, structure, and futures given current state. Despite applicable research in deep filtering and sensor fusion approaches \cite{Krishnan2015,Jain2016}, likely making predictions in the original sensory-space holds promise in robustness for planning algorithms, since actions are directly derived from future scenes. Many encoder-decoder networks hope to achieve this by upsampling to generate predictions in visual (more specifically, sensory) space. Unfortunately, the prediction in visual space may be nonsense. For instance, recall many of these hill-climbing algorithms to fool the network into predicting visually inplausible images. As such, a particular line of work have looked into guarantees that the prediction has physical properties that are relevant and meaningful in real life. \cite{Goodfellow2014} proposed the use of \emph{generative adversarial networks} (GANs) that have been widely accepted as a tool to generate visually plausible predictions that fall into the realm of reality, rather than blurred or meaningless output. This is accomplished by simulating two models: a discriminative model $D$ and a generative model $G$ who is trained to maximize the probability of $D$ making a mistake---this framework is analogous to a minimax two-player game. Building off this work, \cite{Mathieu2015} incorporated the adversarial training techniques into their convolutional network architectures to deal with blur resulting from standard mean squared error loss. They showed that their network was capable of predicting vivid future scenes under an image gradient difference loss function given a set of input sequences. 
	
	Learning the transition dynamics of entities in the world has immediate correlation with an understanding of  intuitive physics. Work by \cite{Lerer2016} incorporated a variant of the \emph{DeepMask} network \cite{Pinheiro2016} that was altered to support multi-class predictions and replicated a number of times to predict the segmentation trajectory of multiple time steps in the future for a falling block prediction task. A ResNet-34 was trained as the trunk of the convolutional network and their approach, \emph{PhysNet}, was shown to outperform all other methods for predicting the future locations of the falling blocks. To insert spatial invariance to neural networks, work by \cite{Jaderberg2015}, introduced a differentiable \emph{Deep Spatial Transformer} module that can be applied to convolutional networks allowing it to be able to explicitly actively transform inherent feature maps. 
	
	\cite{Jain2015} presented a generic framework to model time-space interactions using statio-temporal graphs with a recurrent neural network architecture. The use of spatio-temporal structures impose high-level intuitions allow for improvements in modeling human motion and predicting object interactions. Prediction work by \cite{Oh2015} has made several interesting discoveries in predicting futures in the gameplay domain (namely in the game Space Invaders). They showed that a feedforward network was better at predicting precise movements of objects when recurrent structures consistently made a few pixels of translation error. Their hypothesis is due to the failure of precise spatio-temporal encodings in the recurrent setting, however, they found that recurrent structures were better at predicting events that have long-term dependencies. Long-term sequential movements of objects as a result of an applied force vector at a particular location in the image were learned by a deep neural network while taking into account the geometry and appearance of the scene by using convolutions and recurrent layers in the network \cite{Mottaghi2016}. Others looked into building models for action-conditioned video prediction that explicitly models the motion of pixels rather than predicting the future as a whole. This is achieved by predicting distributions over pixel motion from previous frames and as a result, the model is partially invariant to occlusions. Their model was trained on a dataset of $50,000$ robot interaction videos and resulted in the learning of a ``visual imagination''---a concept of predicting different futures based on the the robot's courses of actions \cite{Finn2016b}. Similarily, \cite{Santana2016} trained a realistic, action-conditioned vehicle simulator using generative adversarial networks. These video prediction mechanisms share a similar action-conditioned form of function approximation as \emph{aspect transition networks} given by $f_{W}: s, \rho_{\phi_i|^\sigma_\tau} \mapsto s'$.
	
	Further work presented by \cite{Agrawal2016} allowed a robot to gather over $400$ hours of experience by poking different objects over $50,000$ times. They learned both an inverse and forward model of the dynamics---the inverse model provided supervision to build informative visual features, which then was used by the forward model to predict the interaction outcomes. They refer to these accurate models for multi-step decision making.

	\subsection{Deriving Environmental Reward}
	\label{subsec:reward}
	
	A likely candidate for environmental reward is one that is computed through an information theoretic interpretation of the affordance prediction networks corresponding to the system's understanding of interaction and dynamics of the world. Assume the robot interacts with the world by executing some set of control programs and obtains interaction tuples $\langle s, \Phi, P, S'\rangle$ describing the initial state $s$, the control programs that were executed $\phi_i|^\sigma_\tau \in \Phi$ with parameters $\rho_i \in P$ resulting in future states $s_i' \in S'$---in turn this outlines some experience dataset described by $\mathcal{D}_t = \{e_1, e_2, \cdots, e_n\}$ where each experience tuple consisting of $e_i = \langle s, \phi_k|^\sigma_\tau, \rho_k, s'\rangle$. 
	
	Now, consider the class of reward structures that adhere to using uncertainty and degree of understanding to penalize or promote the selection of new actions and behaviors to emerge. A prominent example of structures of this nature is the differential variance intrinsically motivated function originally proposed by \cite{Hart2009}. Such a function motivates systems to perform actions that it is most uncertain about, allow it to exploit this reward to build new behaviors. \cite{Wong-ICDL2016} proposed to use this function to learn complete affordance models by showing that the system consumes rewards as representations become more accurate. Unfortunately, these metrics imply a Gaussian distribution assumption that likely fails when adapting to high dimensional sensory-spaces, as such candidate surrogates are information theoretic functions that prescribe distance metrics on predictions and truths. As such consider, 
	\[I_{f_W} = H(f_W(s, \rho_{\phi_i|^\sigma_\tau}))+H(s')-H(f_W(s, \rho_{\phi_i|^\sigma_\tau}), s')\]
	where, $I_{f_W} = I(f_W(s, \rho_{\phi_i|^\sigma_\tau}); s'))$ expresses the mutual information between the prediction of what the outcome should be according to the network $f_W$ given state $s$ and control parameters $\rho_{\phi_i|^\sigma_\tau}$ and the actual outcome $s'$---in essence, a measure of the similarity between prediction $f_W(s, \rho_{\phi_i|^\sigma_\tau})$ and result $s'$. 
	
	We consider this a candidate reward scheme and express its mechanics under two plausible scenarios, 
	
	\subsubsection*{Scenario 1, Low Mutual Information:} 
	In the case that there is low mutual information between the output of the affordance network $f_W$ and the true outcome $s'$, this is attributed to two likely culprits, either the network approximating $f_W$ is not converged, in which case, the system does not have a good understanding of the underlying dynamics of the world, or the action networks corresponding to $f_{\gamma^T}, f_{\rho}, f_{\pi^*_{N}}$ do not approximate the appropriate control parameters or falsely predicts activations of primitives in the world. Either way, these networks are continuously trained with current dataset $\mathcal{D}_t$. 
	
	\subsubsection*{Scenario 2, High Mutual Information:}
	In the case that there is high mutual information between these quantities, the system has acquired good approximations to interaction outcomes via $f_W$ given its current set of controllers expressed as both primitives $\phi_i|^\sigma_\tau \in \Phi$ and complex encodings $\pi^*_j \in \Pi^*$. And those behaviors have likely found useful control goals describe by the approximator $f_{\rho_i}$. As such, this becomes a state of either habituation or emergent behavior---when high mutual information is observed, a likely course of action to continue the development of complex behaviors and interaction-based data collection is to spawn a new network $f_{\pi^*_{n+1}}$ with the sole purpose of attempting to learn new behavioral control sequences and compositions that result in unexpected transition dynamics in the world. Simply, the control policy is rewarded when it learns sequences of actions that fool the dynamics prediction network $f_W$ into new states, while the $f_W$ networks uses these novel interactions to refine its inherent representation. 
	
	\mbox{}\\
	\indent This direction of thinking is adapted from a form of adversarial training where the affordance network tries to best predict the outcome of futures under interactions while the behavioral networks attempts to learn new control parameters that the affordance network fails to predict---hence, new behaviors develop that broaden the system's understanding of the world. From this fact, the reward given to the predictor $f_W$ and the reward given to the developing behavioral policy network $f_{\pi^*_{n+1}}$ can not be and should not be the same. In fact, they exhibit an inverse variation phenomena, therefore, the reward for new behavioral policy networks must be derived from the prediction network's ill-performance. An example reward structure that obeys this particular property is the Kullback-Leibler (KL) divergence given by, $\mathcal{D}(f_W(s, \rho_{\phi_k|^\sigma_\tau})||s') = \sum_{i,j} f_W(s, \rho_{\phi_k|^\sigma_\tau})_{i,j} \log(f_W(s, \rho_{\phi_k|^\sigma_\tau})_{i,j}/s_{i,j}$
	---this particular quantity is ubiquitously used as a measure of the similarity between two distributions, describing relative entropy. Such metrics generally concern controlling the exploration and exploitation tradeoffs in learning architectures \cite{Levine2015}. For instance, a number of approaches have used the KL-divergence to control policy update step sizes \cite{Peters2010,Levine2014,Schulman2015,Akrour2016}. 
	
	A key concern with this approach  is the stability of the learned control policy under an ever-changing dynamics model may be compromised, especially when computing rewards in response to its predictive accuracy. To address this, it is wise to consider a  target network paradigm like that of \cite{Mnih2015}, except instead of freezing the target $Q$ network for stability, one should consider freezing the affordance network $f_W$ when computing rewards for the corresponding behavioral policy network. Since otherwise, these rewards will be drastically non-stationary and thus may have large implications on convergence issues.

	\section{Implications for Task Planning} 
	\label{sec:planning}
	Perhaps, one of the most recurrent themes throughout this manuscript is that there likely is not a single method capable of solving all problems, especially those that concern developing artificial intelligence for physical robotic systems. Similarly, we find that a candidate approach is the marriage between several \emph{theisms} of thought. For instance, consider a lifelong learning framework that generates useful artifacts that task planners can exploit---in actuality, let us briefly entertain this idea. Quite obviously, we would like robots to perform useful tasks or missions in the real world given its massive repertoire of motor skills and precise, learned representation of interaction dynamics in the world. Because these representations and control policies are all derived by the robot through intrinsic motivation, it encodes inherent uncertainties that allow for robust plans and execution of actions. So evidently, it is up to task planners to find plans over these control skills and transition dynamics such that the robot will solve useful problems\footnote{While a complete review of all task planning frameworks is outside the scope of this paper, we discuss how lifelong learning artifacts can integrate with a number of selected planners to accomplish task-relevant solutions.}. Planning generally assumes some form of forward dynamics model, of how actions affect the state of the world---in this particular scenario, we suggest that a learned aspect transition network $f_W$ will serve purposefully as it is a representation for forward dynamics. In other works, basic push motion planning was achieved using dynamics in the form of video prediction and visual foresight \cite{Finn2016c}. In another study by \cite{Tamar2016}, \emph{value iteration networks} were presented as an architecture that allows systems with the capability of \emph{learning to plan} by embedding the fully differentiable neural network with a ``planning module.'' 
	
	
	
	
	Interestingly, research has  shown that an aspect geometry alone is sufficient in describing a number of complex robotic tasks \cite{Ruiken2016b}. In particular, object identification and assembly tasks can be reconfigured into a model-referenced belief-space planner. The aspect definition prescribes sensory geometries to define a Markovian state described by an aspect node in a geometric structure outlining the geometric constellations under some field of view---thus, encoding latent affordances of entities in the world. The specific geometries of features embedded in the environment can be used to drive belief-space architectures into task-specific solutions. Simply in this setting, the artifacts produced by the approximations describe the networks $f_{\gamma^T}, f_{\pi^*_{}}, f_{\rho}$ and $f_W$, which are respectively the networks for control state prediction, complex behavioral policies, continuous control parameters, and world transition dynamics, can be used during planning rollouts (i.e. a Monte Carlo simulation of trajectories according to action-conditioned transition dynamics $f_W$ via parameters $f_\rho$). 
	
	We decompose this task solution into a mathematical representation outlined by a (partially observable) Markov Decision Process. Firstly, Markovian state is given by the robot's perception or raw sensory input. The set of actions in this case collectively describe the set of all control state predictors, control parameter learning networks, and complex policies given by the three set of networks: $f_{\gamma^T}, f_{\pi^*}, f_{\rho}$. And lastly, the transition dynamics are encoded through the aspect transition network $f_W$, with actions being constrained through the $\gamma^T$-networks' prediction that decides the likely control states at any given time instance. 
	
	In planning, one may simply perform rollouts over the Markovian state and predict likely candidate actions that can be executed. Many planning approaches at this point resort to sample techniques in conjunction with RRTs, preimage backchaining \cite{Kaelbling2011}, or large population of dynamics models \cite{Ruiken2016}. Instead, we have control state networks that specifically describe candidate control programs---for each of these actions, we rollout under the aspect transition network the candidate future states. A similar affordance model referenced \emph{Active Belief Planner} \cite{Ruiken2016} has been presented recently to solve object identification tasks. In fact, their planner uses these affordance representations as forward models during problem solving behavior in non-ideal contexts that include sensor noise, suboptimal lighting, missing information, and extraneous information arising from scenes that can contain multiple objects in initially unknown arrangements. Unfortunately, their methods requires that large populations of transition models are  explicitly described---this population is what would be a useful structure to approximate using $f_W$, the aspect transition network. 
	
	Evidently this roll out can be performed by a number of generic planners that expand states accordingly to their future dynamics. As such, a number of planners like A*, All-Domain Execution and Planning Technology (ADEPT) \cite{Ricard2003}, or Hierarchical Planning in the Now (HPN) \cite{Kaelbling2011} can solve for task relevant plans while leveraging learned artifacts for additional robustness. 
	
	A fundamental problem with planners in general consist of the planning horizon and the branching factor prescribed by the possible number of actions at any given state. Hierarchy attempts to reduce the planning horizon by only planning to the first executable primitive. Still, hierarchical planning scales exponentially (number of operators to the shallowest abstraction level). Fortunately, the control state description networks $f_{\gamma^T}$ and complex behavioral policy networks $f_{\pi^*}$ drastically helps reduce this planning complexity by in turn decreasing the depth at which the planner must roll out due to the consideration of more complex motions. Secondly, the possible actions at any given state is constrained by those physically plausible which is immediately evaluated by the approximator $f_{\gamma^T_i}$ for each control program $\phi_i|^\sigma_\tau$---quiescent actions should then not be considered. 
	
	Interestingly, another problem that arises is when a cognitive system increases its skill set, this in turn has monotonically increasing effects on the branching factor of planners. One may consider only feasible transitions to attack this phenomena---the feasibility of actions requires either geometric evaluations or explicitly defined dynamics models. For instance, HPN uses generators that reason over the geometry of entities in the world \cite{Kaelbling2011}. The concept of an aspect transition graph under planning frameworks do a good job of ensuring that only feasible actions are planned for by exploiting the likely transitions under learned object models \cite{Ruiken2016}. Similarily, the network variants are good candidates for quickly evaluating potential control contexts and feasibility of particular actions parameters. 
	
	
	
	Importantly, by no means do the control policies and transition dynamics learned through lifelong intrinsic motivation limit the use of more sophisticated planners. Dealing with unstructured and dynamic environments becomes a fundamental problem. Therein, a number of frameworks have been adjusted to operate over uncertainties like for example, the Active Belief Planner \cite{Ruiken2016} and the Belief-space HPN \cite{Kaelbling2013}. 
	
	A key insight is that many of these video prediction paradigms (as described in Section~\ref{sec:affordance}) can be inferred as affordance models that predict all possible futures given interaction with the world---a promising advantage of using generative adversarial networks. As such, this inherently encodes the belief over many possible outcomes that may result in interaction and can be leveraged in planning. For instance, one can operate in the belief space of futures by observing the manifold on which the many futures lie. It appears that deep generative adversarial networks, like the future prediction networks trained adversarially, obey certain arithmetics \cite{Radford2015} and as a result can be used to discover such a futures manifold.

	
	
	\section{Conclusion}
	This paper has provided an initial survey of recent advances in deep learning applicable to mobile perceptual systems, namely pertaining to the robotics domain. We discuss a series of challenges that arise when applied to physical embodied systems that are otherwise unseen in strictly vision and simulation domains. And we have outlined these recent advances in detection, control, and future prediction problems that are most relevant to robotics and candidate planners in a new learning direction. 
	
	These advances were structured in this manuscript in such a way that implies a future direction in self-supervision and lifelong learning. Piecing together these individual research ideas, we indicate that the technologies currently may be ripe to design a lifelong self-supervised system to learn complex behaviors in the real world. As such, the acquisition over an extended period of learning can be leveraged in numerous robotics tasks both in research and industry by coupling these learned artifacts with existing task planners. As \cite{Silver2013} mentions, it is time to move on from task-specific machine learning---instead, learn over an extensive repertoire, over numerous tasks in order to acquire general intelligence. 
	
	However, with using physical hardware a concern revolving around exploration of control actions comes into play. One of the most fundamental concerns with learning systems considers the question of what constitutes a \emph{safe} exploration paradigm. Rather, how does one ensure that the system does not perform catastrophic actions during exploration? Safe reinforcement paradigms have been outlined by \cite{Thomas-Thesis2015}, discussing algorithms to search for new and refined policies while ensuring that the probability of bad policies are minimized. In these works, \cite{Thomas2015} presented a method using the trajectories of other policies that were executed in the past to efficiently, with high confidence, perform off-policy evaluations to gauge exploration candidates. With such an approach, it becomes possible to evaluate the performance of new policies without explicit execution. Perhaps, the future for self-supervised systems lies in the connection to metrics that safeguards the hardware while effectively evaluating its possible actions. Measures like these should be considered in order to built a system that learns over a lifetime of experiences. 
	
	Recently, the emergence of \emph{deep symbolic reinforcement learning} may be a promising architecture by combining recent breakthroughs in deep reinforcement learning with classical symbolic artificial intelligence \cite{Garnelo2016}. Simply, a neural network backend is used to extract useful symbolic representations which are then used by a symbolic frontend for action selection. Although the work is still at its infancy, a fundamental drawback is that alike the formulation of DQN, it requires that the system has a specified task that it is trying to solve in which reward can be evaluated from. In fact, the resulting artifact of this is a meta-policy composed of sub-policies under a specified task---these sub-policies are however locally optimal under any combination of interactions between entities. There are two evident issues consisting of scaling, due to the nature of considering all possible interactions, and troubles with global optima. However, insight from the two network approach, learning a value and a policy network, may help support some of these immediate issues. Perhaps as this idea develops, it may be considered as a module in lifelong self-supervision, due to its promising connections with symbolic hierarchical planning \cite{Kaelbling2013}.   
	
	An important aspect of reinforcement learning is the capability of transfer, both between systems and between task domains. Work by \cite{Devin2016} provided insight on the decomposition of network policies into robot-specific and task-specific modules that supported transfer between tasks and different robot morphologies (e.g. varying in number of links and joints). Interestingly, under the \emph{control basis formulation}, parametrized controllers already supports generalization from robot to robot, with an assumption that the new system has sufficiently motor resources the same control objectives. As such, the high level behavioral networks, those that are composed of primitive parametrized controllers, too inherit this form of generalizability. 
	
	A good review by \cite{Lake2016} discusses the fundamental cognitive problems with building systems that expertly accomplish tasks by pattern recognition alone. Wherein to build cognitive systems that learn and think like people, they suggest that these systems must have the capability to support both explanation and understanding. Systems must be able to understand intuitive physics, have the capacity to learn to learn, and build grounded generalizations that span new tasks and situations---such a view is similar to the direction presented in this paper. Our survey is particularly tailored towards the connection of these ideas with physical robot systems. We suggest that perhaps the goal is not the build a system that exactly mimics human cognition and learning, but instead draw \emph{insight} and \emph{computational analogs} from ideas in cognitive development. Robots are not humans and do not necessarily have to learn at their granularity nor produce the same artifacts through learning. But, we regard that studying the development of cognition in biological systems may be crucial in building algorithms for artificial systems. 
	
	In this manuscript, we outlined powerful nonlinear approximation tools with inspirations from cognitive development and control theory to produce a direction in which lifelong learning frameworks can be applied to autonomous systems that continuously acquire a hierarchy of complex motor behaviors in addition to a dynamics representation of interactions in the world. A number the ideas presented in this manuscript were influenced by the computational development of action and representation going back to \cite{Grupen2005}. Their work, however, assumes there exists some preordained developmental guideline under the notion of a \emph{Developmental Assembler} that provides design constraints and developmental schedules. Such entities assigns task specific rewards that are the fundamental motivators to the acquisition of complex behaviors. In the case of our review, we outlined an example reward paradigm that computes reward that adhere to the system's internal predictions of the world and of it evolution through interaction---tying together the dynamic modeling of action related complexes in the world. Further, we surveyed numerous deep learning advances pertaining to robotics and found a close connection between many deep reinforcement learning paradigms with classical concurrent control schemes under the \emph{control basis} formulation. With this survey, we would like to acknowledge that deep learning should be considered as an hyperparametric approximation tool that alone is likely not capable of attacking all problems. And as such, we foresee a direction in which these powerful approximators are integrated with closed-loop control and optimization paradigms, driven by principled motivators, and inspired by insight from cognitive development to realize a robust lifelong self-supervised system.

	\section*{Acknowledgements}
	Although the author's affiliations are with the Charles Stark Draper Laboratory, Inc, any opinions, findings, conclusions, or recommendations expressed in this material are solely those of the author(s) and do not necessarily reflect the views of organization. 
	
	I would like to thank Takeshi Takahashi and Roderic A. Grupen for their expertise, enthusiasm, and insight on preliminary work with control state prediction networks that lead up to this direction of thinking. 
	
	Lastly, and most importantly, the stand, the structure, and the direction of this manuscript would not be in its current form, without the critiques and ``skepticisms'' of my good friend Mitchell Hebert. 
	

	\addcontentsline{toc}{section}{{\bf References}} 
	\bibliographystyle{SageH}      
	
	\bibliography{developmental,deep-learning,grupen-master,vc,robots}

\begin{thebibliography}{211}
\providecommand{\natexlab}[1]{#1}
\providecommand{\url}[1]{\texttt{#1}}
\providecommand{\urlprefix}{URL }
\expandafter\ifx\csname urlstyle\endcsname\relax
  \providecommand{\doi}[1]{DOI:\discretionary{}{}{}#1}\else
  \providecommand{\doi}{DOI:\discretionary{}{}{}\begingroup
  \urlstyle{rm}\Url}\fi

\bibitem[{Agrawal et~al.(2016)Agrawal, Nair, Abbeel, Malik and
  Levine}]{Agrawal2016}
Agrawal P, Nair A, Abbeel P, Malik J and Levine S (2016) Learning to poke by
  poking: Experiential learning of intuitive physics.
\newblock \emph{arXiv preprint:1606.07419} .

\bibitem[{Akrour et~al.(2016)Akrour, Abdolmaleki, Abdulsamad and
  Neumann}]{Akrour2016}
Akrour R, Abdolmaleki A, Abdulsamad H and Neumann G (2016) Model-free
  trajectory optimization for reinforcement learning.
\newblock \emph{arXiv preprint:1606.09197} .

\bibitem[{Aronson(1981)}]{Aronson81}
Aronson A (1981) \emph{Clinical Examinations in Neurology}.
\newblock Philadelphia, PA: W.B. Saunders Co.

\bibitem[{Arulkumaran et~al.(2016)Arulkumaran, Dilokthanakul, Shanahan and
  Bharath}]{Arulkumaran2016}
Arulkumaran K, Dilokthanakul N, Shanahan M and Bharath AA (2016) Classifying
  options for deep reinforcement learning.
\newblock \emph{arXiv preprint:1604.08153} .

\bibitem[{Badrinarayanan et~al.(2015)Badrinarayanan, Kendall and
  Cipolla}]{Badrinarayanan2015}
Badrinarayanan V, Kendall A and Cipolla R (2015) Segnet: A deep convolutional
  encoder-decoder architecture for image segmentation.
\newblock \emph{arXiv preprint:1511.00561} .

\bibitem[{Baird~III(1993)}]{Baird1993}
Baird~III LC (1993) Advantage updating.
\newblock Technical report, DTIC Document.

\bibitem[{Balduzzi and Ghifary(2015)}]{Balduzzi2015}
Balduzzi D and Ghifary M (2015) Compatible value gradients for reinforcement
  learning of continuous deep policies.
\newblock \emph{arXiv preprint:1509.03005} .

\bibitem[{Barto(2013)}]{Barto2013}
Barto AG (2013) Intrinsic motivation and reinforcement learning.
\newblock In: \emph{Intrinsically motivated learning in natural and artificial
  systems}. Springer, pp. 17--47.

\bibitem[{Barto et~al.(2004)Barto, Singh and Chentanez}]{Barto2004}
Barto AG, Singh S and Chentanez N (2004) {Intrinsically Motivated Learning of
  Hierarchical Collections of Skills}.
\newblock In: \emph{International Conference on Developmental Learning}.

\bibitem[{Bendale and Boult(2015)}]{Bendale2015}
Bendale A and Boult T (2015) Towards open set deep networks.
\newblock \emph{arXiv preprint:1511.06233} .

\bibitem[{Bengio(2009)}]{Bengio2009}
Bengio Y (2009) Learning deep architectures for ai.
\newblock \emph{Foundations and trends in Machine Learning} 2(1): 1--127.

\bibitem[{Bojarski et~al.(2016)Bojarski, Del~Testa, Dworakowski, Firner, Flepp,
  Goyal, Jackel, Monfort, Muller, Zhang et~al.}]{Bojarski2016}
Bojarski M, Del~Testa D, Dworakowski D, Firner B, Flepp B, Goyal P, Jackel LD,
  Monfort M, Muller U, Zhang J et~al. (2016) End to end learning for
  self-driving cars.
\newblock \emph{arXiv preprint:1604.07316} .

\bibitem[{Brooks(1987{\natexlab{a}})}]{Brooks-FAI87}
Brooks R (1987{\natexlab{a}}) Intelligence without representation.
\newblock In: \emph{Proceedings of the Workshop on the Foundations of
  Artificial Intelligence}. MIT Press.

\bibitem[{Brooks(1987{\natexlab{b}})}]{Brooks1987}
Brooks RA (1987{\natexlab{b}}) Planning is just a way of avoiding figuring out
  what to do next .

\bibitem[{Byravan and Fox(2016)}]{Byravan2016}
Byravan A and Fox D (2016) Se3-nets: Learning rigid body motion using deep
  neural networks.
\newblock \emph{arXiv preprint:1606.02378} .

\bibitem[{Carlini and Wagner(2016)}]{Carlini2016}
Carlini N and Wagner D (2016) Towards evaluating the robustness of neural
  networks.
\newblock \emph{arXiv preprint:1608.04644} .

\bibitem[{Chebotar et~al.(2016)Chebotar, Kalakrishnan, Yahya, Li, Schaal and
  Levine}]{Chebotar2016}
Chebotar Y, Kalakrishnan M, Yahya A, Li A, Schaal S and Levine S (2016) Path
  integral guided policy search.
\newblock \emph{arXiv preprint:1610.00529} .

\bibitem[{Chemero(2003)}]{Chemero03}
Chemero A (2003) An outline of a theory of affordances.
\newblock \emph{Ecological Psychology} 15(3): 181--195.

\bibitem[{Coelho(2001)}]{Coelho-Thesis01}
Coelho J (2001) \emph{Multifingered Grasping: Haptic Reflexes and Control
  Context}.
\newblock PhD Thesis, University of Massachusetts, Amherst, MA.

\bibitem[{Coelho and Grupen(1997)}]{Coelho_JRS97}
Coelho J and Grupen R (1997) A control basis for learning multifingered grasps.
\newblock \emph{Journal of Robotic Systems} 14(7): 545--557.

\bibitem[{Coelho and Grupen(1998)}]{Coelho1998}
Coelho JA and Grupen R (1998) Dynamic control models as state abstractions.
\newblock In: \emph{Neural Information Processing Systems Workshop: Abstraction
  and Hierarchy in Reinforcement Learning}.

\bibitem[{Connolly and Grupen(1993)}]{Connolly1993}
Connolly CI and Grupen RA (1993) The applications of harmonic functions to
  robotics.
\newblock \emph{Journal of Robotic Systems} 10(7): 931--946.

\bibitem[{Da~Silva et~al.(2012)Da~Silva, Konidaris and Barto}]{DaSilva2012}
Da~Silva B, Konidaris G and Barto A (2012) Learning parameterized skills.
\newblock \emph{arXiv preprint:1206.6398} .

\bibitem[{Dellin et~al.(2016)Dellin, Strabala, Haynes, Stager and
  Srinivasa}]{Dellin2016}
Dellin CM, Strabala K, Haynes GC, Stager D and Srinivasa SS (2016) Guided
  manipulation planning at the darpa robotics challenge trials.
\newblock In: \emph{Experimental Robotics}. Springer, pp. 149--163.

\bibitem[{Devin et~al.(2016)Devin, Gupta, Darrell, Abbeel and
  Levine}]{Devin2016}
Devin C, Gupta A, Darrell T, Abbeel P and Levine S (2016) Learning modular
  neural network policies for multi-task and multi-robot transfer.
\newblock \emph{arXiv preprint:1609.07088} .

\bibitem[{Feng et~al.(2015{\natexlab{a}})Feng, Whitman, Xinjilefu and
  Atkeson}]{Feng2015}
Feng S, Whitman E, Xinjilefu X and Atkeson CG (2015{\natexlab{a}})
  Optimization-based full body control for the darpa robotics challenge.
\newblock \emph{Journal of Field Robotics} 32(2): 293--312.

\bibitem[{Feng et~al.(2015{\natexlab{b}})Feng, Xinjilefu, Atkeson and
  Kim}]{Feng2015b}
Feng S, Xinjilefu X, Atkeson CG and Kim J (2015{\natexlab{b}}) Optimization
  based controller design and implementation for the atlas robot in the darpa
  robotics challenge finals.
\newblock In: \emph{IEEE-RAS 15th International Conference on Humanoid Robots
  (Humanoids)}. pp. 1028--1035.

\bibitem[{Finn et~al.(2016{\natexlab{a}})Finn, Goodfellow and
  Levine}]{Finn2016b}
Finn C, Goodfellow I and Levine S (2016{\natexlab{a}}) Unsupervised learning
  for physical interaction through video prediction.
\newblock \emph{arXiv preprint:1605.07157} .

\bibitem[{Finn and Levine(2016)}]{Finn2016c}
Finn C and Levine S (2016) Deep visual foresight for planning robot motion.
\newblock \emph{arXiv preprint:1610.00696} .

\bibitem[{Finn et~al.(2016{\natexlab{b}})Finn, Levine and Abbeel}]{Finn2016a}
Finn C, Levine S and Abbeel P (2016{\natexlab{b}}) Guided cost learning: Deep
  inverse optimal control via policy optimization.
\newblock \emph{arXiv preprint:1603.00448} .

\bibitem[{Finn et~al.(2015)Finn, Tan, Duan, Darrell, Levine and
  Abbeel}]{Finn2015}
Finn C, Tan XY, Duan Y, Darrell T, Levine S and Abbeel P (2015) Deep spatial
  autoencoders for visuomotor learning.
\newblock \emph{arXiv preprint:1509.06113} .

\bibitem[{Frank et~al.(2015)Frank, Leitner, Stollenga, F{\"o}rster and
  Schmidhuber}]{Frank2015}
Frank M, Leitner J, Stollenga M, F{\"o}rster A and Schmidhuber J (2015)
  Curiosity driven reinforcement learning for motion planning on humanoids.
\newblock \emph{Intrinsic motivations and open-ended development in animals,
  humans, and robots} .

\bibitem[{{Garnelo} et~al.(2016){Garnelo}, {Arulkumaran} and
  {Shanahan}}]{Garnelo2016}
{Garnelo} M, {Arulkumaran} K and {Shanahan} M (2016) {Towards Deep Symbolic
  Reinforcement Learning}.
\newblock \emph{arXiv preprint:1609.05518} .

\bibitem[{Ge and Cui(2002)}]{Ge2002}
Ge SS and Cui YJ (2002) Dynamic motion planning for mobile robots using
  potential field method.
\newblock \emph{Autonomous Robots} 13(3): 207--222.

\bibitem[{Gibson(1988)}]{Gibson1988}
Gibson EJ (1988) Exploratory behavior in the development of perceiving, acting,
  and the acquiring of knowledge.
\newblock \emph{Annual review of psychology} 39(1): 1--42.

\bibitem[{Gibson(1977)}]{GibsonJJ77}
Gibson JJ (1977) The theory of affordances.
\newblock In: Shaw R and Bransford J (eds.) \emph{Perceiving, Acting, and
  Knowing: Toward an Ecological Psychology}, chapter~3. Hillsdale, New Jersey:
  Lawrence Erlbaum, pp. 67--82.

\bibitem[{Goodfellow et~al.(2016)Goodfellow, Bengio and
  Courville}]{Goodfellow2016}
Goodfellow I, Bengio Y and Courville A (2016) Deep learning.
\newblock \emph{Book in preparation for MIT Press} .

\bibitem[{Goodfellow et~al.(2014{\natexlab{a}})Goodfellow, Pouget-Abadie,
  Mirza, Xu, Warde-Farley, Ozair, Courville and Bengio}]{Goodfellow2014}
Goodfellow I, Pouget-Abadie J, Mirza M, Xu B, Warde-Farley D, Ozair S,
  Courville A and Bengio Y (2014{\natexlab{a}}) Generative adversarial nets.
\newblock In: \emph{Advances in Neural Information Processing Systems}. pp.
  2672--2680.

\bibitem[{Goodfellow et~al.(2014{\natexlab{b}})Goodfellow, Shlens and
  Szegedy}]{Goodfellow2015a}
Goodfellow IJ, Shlens J and Szegedy C (2014{\natexlab{b}}) Explaining and
  harnessing adversarial examples.
\newblock \emph{arXiv preprint:1412.6572} .

\bibitem[{Grupen and Huber(2005)}]{Grupen2005}
Grupen RA and Huber M (2005) A framework for the development of robot behavior.
\newblock \emph{AAAI Spring Symposium on Developmental Robotics} .

\bibitem[{Gu et~al.(2016{\natexlab{a}})Gu, Holly, Lillicrap and
  Levine}]{Gu2016b}
Gu S, Holly E, Lillicrap T and Levine S (2016{\natexlab{a}}) Deep reinforcement
  learning for robotic manipulation.
\newblock \emph{arXiv preprint:1610.00633} .

\bibitem[{Gu et~al.(2016{\natexlab{b}})Gu, Lillicrap, Sutskever and
  Levine}]{Gu2016}
Gu S, Lillicrap T, Sutskever I and Levine S (2016{\natexlab{b}}) Continuous
  deep q-learning with model-based acceleration.
\newblock \emph{arXiv preprint:1603.00748} .

\bibitem[{Gullapalli et~al.(1992)Gullapalli, Grupen and Barto}]{Gullapalli1992}
Gullapalli V, Grupen RA and Barto AG (1992) Learning reactive admittance
  control.
\newblock In: \emph{IEEE International Conference on Robotics and Automation}.
  pp. 1475--1480.

\bibitem[{Hariharan et~al.(2015)Hariharan, Arbel{\'a}ez, Girshick and
  Malik}]{Hariharan2015}
Hariharan B, Arbel{\'a}ez P, Girshick R and Malik J (2015) Hypercolumns for
  object segmentation and fine-grained localization.
\newblock In: \emph{Proceedings of the IEEE Conference on Computer Vision and
  Pattern Recognition}. pp. 447--456.

\bibitem[{Harmon and Baird~III(1996)}]{Harmon1996}
Harmon ME and Baird~III LC (1996) Multi-player residual advantage learning with
  general function approximation.
\newblock \emph{Wright Laboratory, WL/AACF, Wright-Patterson Air Force Base,
  OH} : 45433--7308.

\bibitem[{Hart(2009{\natexlab{a}})}]{Hart2009thesis}
Hart S (2009{\natexlab{a}}) \emph{The Development of Hierarchical Knowledge in
  Robot Systems}.
\newblock PhD Thesis, University of Massachusetts Amherst.

\bibitem[{Hart(2009{\natexlab{b}})}]{Hart2009}
Hart S (2009{\natexlab{b}}) An intrinsic reward for affordance exploration.
\newblock In: \emph{IEEE International Conference on Development and Learning}.

\bibitem[{Hausknecht and Stone(2015)}]{Hausknecht2015}
Hausknecht MJ and Stone P (2015) Deep reinforcement learning in parameterized
  action space.
\newblock \emph{CoRR} arXiv preprint:1511.04143.

\bibitem[{Hebert et~al.(2015)Hebert, Wong and Grupen}]{Hebert2015}
Hebert M, Wong JM and Grupen RA (2015) Phase lag bounded velocity planning for
  high performance path tracking.
\newblock In: \emph{IEEE-RAS 15th International Conference on Humanoid Robots}.
  pp. 947--952.

\bibitem[{Heess et~al.(2015{\natexlab{a}})Heess, Hunt, Lillicrap and
  Silver}]{Heess2015}
Heess N, Hunt JJ, Lillicrap TP and Silver D (2015{\natexlab{a}}) Memory-based
  control with recurrent neural networks.
\newblock \emph{arXiv preprint:1512.04455} .

\bibitem[{Heess et~al.(2015{\natexlab{b}})Heess, Wayne, Silver, Lillicrap, Erez
  and Tassa}]{Heess2015b}
Heess N, Wayne G, Silver D, Lillicrap T, Erez T and Tassa Y
  (2015{\natexlab{b}}) Learning continuous control policies by stochastic value
  gradients.
\newblock In: \emph{Advances in Neural Information Processing Systems}. pp.
  2944--2952.

\bibitem[{Heess et~al.(2016)Heess, Wayne, Tassa, Lillicrap, Riedmiller and
  Silver}]{Heess2016}
Heess N, Wayne G, Tassa Y, Lillicrap T, Riedmiller M and Silver D (2016)
  Learning and transfer of modulated locomotor controllers.
\newblock \emph{arXiv preprint:1610.05182} .

\bibitem[{Hemion(2016{\natexlab{a}})}]{Hemion2016b}
Hemion NJ (2016{\natexlab{a}}) Context discovery for model learning in
  partially observable environments.
\newblock \emph{arXiv preprint:1608.00737} .

\bibitem[{Hemion(2016{\natexlab{b}})}]{Hemion2016}
Hemion NJ (2016{\natexlab{b}}) Discovering latent states for model learning:
  Applying sensorimotor contingencies theory and predictive processing to model
  context.
\newblock \emph{arXiv preprint:1608.00359} .

\bibitem[{Hester and Stone(2015)}]{Hester2015}
Hester T and Stone P (2015) Intrinsically motivated model learning for
  developing curious robots.
\newblock \emph{Artificial Intelligence} .

\bibitem[{Hornik et~al.(1989)Hornik, Stinchcombe and White}]{Hornik1989}
Hornik K, Stinchcombe M and White H (1989) Multilayer feedforward networks are
  universal approximators.
\newblock \emph{Neural networks} 2(5): 359--366.

\bibitem[{Huber and Grupen(1997)}]{Huber-IJCAI97}
Huber M and Grupen R (1997) Learning to coordinate controllers - reinforcement
  learning on a control basis.
\newblock In: \emph{Proceedings of the Fifteenth International Joint Conference
  on Artificial Intelligence (IJCAI)}. Nagoya, JP: IJCAII.

\bibitem[{Huber et~al.(1996{\natexlab{a}})Huber, MacDonald and
  Grupen}]{Huber-ICRA96}
Huber M, MacDonald W and Grupen R (1996{\natexlab{a}}) A control basis for
  multilegged walking.
\newblock In: \emph{Proceedings of the Conference on Robotics and Automation}.
  Minneapolis, MN: IEEE.

\bibitem[{Huber et~al.(1996{\natexlab{b}})Huber, MacDonald and
  Grupen}]{Huber1996}
Huber M, MacDonald WS and Grupen RA (1996{\natexlab{b}}) A control basis for
  multilegged walking.
\newblock In: \emph{IEEE International Conference on Robotics and Automation}.

\bibitem[{Hussein et~al.(2016)Hussein, Gaber and Elyan}]{Hussein2016}
Hussein A, Gaber MM and Elyan E (2016) Deep active learning for autonomous
  navigation.
\newblock In: \emph{International Conference on Engineering Applications of
  Neural Networks}. Springer, pp. 3--17.

\bibitem[{Ijspeert et~al.(2003)Ijspeert, Nakanishi and
  Schaal}]{Ijspeert_NIPS03}
Ijspeert A, Nakanishi J and Schaal S (2003) Learning attractor landscapes for
  learning motor primitives.
\newblock In: Becker S, Thrun S and Obermayer K (eds.) \emph{Advances in Neural
  Information Processing Systems 15}. MIT Press, pp. 1547--1554.

\bibitem[{Jaderberg et~al.(2015)Jaderberg, Simonyan, Zisserman
  et~al.}]{Jaderberg2015}
Jaderberg M, Simonyan K, Zisserman A et~al. (2015) Spatial transformer
  networks.
\newblock In: \emph{Advances in Neural Information Processing Systems}. pp.
  2017--2025.

\bibitem[{Jain et~al.(2016)Jain, Singh, Koppula, Soh and Saxena}]{Jain2016}
Jain A, Singh A, Koppula HS, Soh S and Saxena A (2016) Recurrent neural
  networks for driver activity anticipation via sensory-fusion architecture.
\newblock In: \emph{IEEE International Conference on Robotics and Automation
  (ICRA)}. pp. 3118--3125.

\bibitem[{Jain et~al.(2015)Jain, Zamir, Savarese and Saxena}]{Jain2015}
Jain A, Zamir AR, Savarese S and Saxena A (2015) Structural-rnn: Deep learning
  on spatio-temporal graphs.
\newblock \emph{arXiv preprint:1511.05298} .

\bibitem[{James and Johns(2016)}]{James2016}
James S and Johns E (2016) 3d simulation for robot arm control with deep
  q-learning.
\newblock \emph{arXiv preprint:1609.03759} .

\bibitem[{Johns et~al.(2016)Johns, Leutenegger and Davison}]{Johns2016}
Johns E, Leutenegger S and Davison AJ (2016) Deep learning a grasp function for
  grasping under gripper pose uncertainty.
\newblock \emph{arXiv preprint:1608.02239} .

\bibitem[{Johnson et~al.(2015)Johnson, Shrewsbury, Bertrand, Wu, Duran, Floyd,
  Abeles, Stephen, Mertins, Lesman et~al.}]{Johnson2015}
Johnson M, Shrewsbury B, Bertrand S, Wu T, Duran D, Floyd M, Abeles P, Stephen
  D, Mertins N, Lesman A et~al. (2015) Team ihmc's lessons learned from the
  darpa robotics challenge trials.
\newblock \emph{Journal of Field Robotics} 32(2): 192--208.

\bibitem[{Kaelbling et~al.(1996)Kaelbling, Littman and Moore}]{Kaelbling1996}
Kaelbling LP, Littman ML and Moore AW (1996) Reinforcement learning: A survey.
\newblock \emph{Journal of artificial intelligence research} 4: 237--285.

\bibitem[{Kaelbling and Lozano-P{\'e}rez(2011)}]{Kaelbling2011}
Kaelbling LP and Lozano-P{\'e}rez T (2011) Hierarchical task and motion
  planning in the now.
\newblock In: \emph{IEEE International Conference on Robotics and Automation
  (ICRA)}. pp. 1470--1477.

\bibitem[{Kaelbling and Lozano-P\'{e}rez(2013)}]{Kaelbling2013}
Kaelbling LP and Lozano-P\'{e}rez T (2013) Integrated task and motion planning
  in belief space.
\newblock \emph{International Journal of Robotics Research} 32(9-10).

\bibitem[{Kardan and Stanley(2016)}]{Kardan2016}
Kardan N and Stanley KO (2016) Fitted learning: Models with awareness of their
  limits.
\newblock \emph{arXiv preprint:1609.02226} .

\bibitem[{Kendall and Cipolla(2015)}]{Kendall2015b}
Kendall A and Cipolla R (2015) Modelling uncertainty in deep learning for
  camera relocalization.
\newblock \emph{arXiv preprint:1509.05909} .

\bibitem[{Kendall et~al.(2015)Kendall, Grimes and Cipolla}]{Kendall2015}
Kendall A, Grimes M and Cipolla R (2015) Posenet: A convolutional network for
  real-time 6-dof camera relocalization.
\newblock In: \emph{Proceedings of the IEEE International Conference on
  Computer Vision}. pp. 2938--2946.

\bibitem[{Khatib(1985)}]{Khatib85}
Khatib O (1985) Real-time obstacle avoidance for manipulators and mobile
  robots.
\newblock In: \emph{ICRA}. IEEE, pp. 500--505.

\bibitem[{Kingma and Welling(2013)}]{Kingma2013}
Kingma DP and Welling M (2013) Auto-encoding variational bayes.
\newblock \emph{arXiv preprint:1312.6114} .

\bibitem[{Kohlbrecher et~al.(2015)Kohlbrecher, Romay, Stumpf, Gupta, Von~Stryk,
  Bacim, Bowman, Goins, Balasubramanian and Conner}]{Kohlbrecher2015}
Kohlbrecher S, Romay A, Stumpf A, Gupta A, Von~Stryk O, Bacim F, Bowman DA,
  Goins A, Balasubramanian R and Conner DC (2015) Human-robot teaming for
  rescue missions: Team vigir's approach to the 2013 darpa robotics challenge
  trials.
\newblock \emph{Journal of Field Robotics} 32(3): 352--377.

\bibitem[{Konidaris and Barto(2009)}]{Konidaris2009}
Konidaris G and Barto AG (2009) Efficient skill learning using abstraction
  selection.
\newblock In: \emph{International Joint Conference on Artificial Intelligence},
  volume~9. pp. 1107--1112.

\bibitem[{Kragic and Christensen(2003)}]{Kragic2003}
Kragic D and Christensen HI (2003) Robust visual servoing.
\newblock \emph{The international journal of robotics research} 22(10-11):
  923--939.

\bibitem[{Krishnan et~al.(2015)Krishnan, Shalit and Sontag}]{Krishnan2015}
Krishnan RG, Shalit U and Sontag D (2015) Deep kalman filters.
\newblock \emph{arXiv preprint:1511.05121} .

\bibitem[{Krizhevsky et~al.(2012)Krizhevsky, Sutskever and
  Hinton}]{Krizhevsky2012}
Krizhevsky A, Sutskever I and Hinton GE (2012) Imagenet classification with
  deep convolutional neural networks.
\newblock In: \emph{Advances in Neural Information Processing Systems}.

\bibitem[{Kruger et~al.(2011)Kruger, Geib, Piater, Petrick, Steedman,
  Worgotter, Ude, Asfour, Kraft, Omrcen, Agostini and Dillmann}]{Kruger2011}
Kruger N, Geib C, Piater J, Petrick R, Steedman M, Worgotter F, Ude A, Asfour
  T, Kraft D, Omrcen D, Agostini A and Dillmann R (2011) Object–action
  complexes: Grounded abstractions of sensory–motor processes.
\newblock \emph{Robotics and Autonomous Systems} 59(10): 740 -- 757.

\bibitem[{Ku et~al.(2014)Ku, Sen, Learned-Miller and Grupen}]{Ku-ECCV14}
Ku L, Sen S, Learned-Miller E and Grupen R (2014) Aspect transition graph: An
  affordance-based model.
\newblock In: \emph{European Conference on Computer Vision, Workshop on
  Affordances: Visual Perception of Affordances and Functional Visual
  Primitives for Scene Analysis}.

\bibitem[{Ku et~al.(2016)Ku, Learned-Miller and Grupen}]{Ku2016}
Ku LY, Learned-Miller E and Grupen R (2016) Associating grasping with
  convolutional neural network features.
\newblock \emph{arXiv preprint:1609.03947} .

\bibitem[{Kuindersma et~al.(2016)Kuindersma, Deits, Fallon, Valenzuela, Dai,
  Permenter, Koolen, Marion and Tedrake}]{Kuindersma2016}
Kuindersma S, Deits R, Fallon M, Valenzuela A, Dai H, Permenter F, Koolen T,
  Marion P and Tedrake R (2016) Optimization-based locomotion planning,
  estimation, and control design for the atlas humanoid robot.
\newblock \emph{Autonomous Robots} 40(3): 429--455.

\bibitem[{Kuindersma et~al.(2009)Kuindersma, Hannigan, Ruiken and
  Grupen}]{Kuindersma09:KnuckleWalking}
Kuindersma S, Hannigan E, Ruiken D and Grupen R (2009) {Dexterous Mobility with
  the uBot-5 Mobile Manipulator}.
\newblock In: \emph{14th International Conference on Advanced Robotics (ICAR)}.
  Munich, Germany.

\bibitem[{Kulkarni et~al.(2016)Kulkarni, Narasimhan, Saeedi and
  Tenenbaum}]{Kulkarni2016}
Kulkarni TD, Narasimhan KR, Saeedi A and Tenenbaum JB (2016) Hierarchical deep
  reinforcement learning: Integrating temporal abstraction and intrinsic
  motivation.
\newblock \emph{arXiv preprint:1604.06057} .

\bibitem[{Kurakin et~al.(2016)Kurakin, Goodfellow and Bengio}]{Kurakin2016}
Kurakin A, Goodfellow I and Bengio S (2016) Adversarial examples in the
  physical world.
\newblock \emph{arXiv preprint:1607.02533} .

\bibitem[{Lake et~al.(2016)Lake, Ullman, Tenenbaum and Gershman}]{Lake2016}
Lake BM, Ullman TD, Tenenbaum JB and Gershman SJ (2016) Building machines that
  learn and think like people.
\newblock \emph{arXiv preprint:1604.00289} .

\bibitem[{Law et~al.(2011)Law, Lee, Hulse and Tomassetti}]{Law2011}
Law J, Lee MH, Hulse M and Tomassetti A (2011) The infant development timeline
  and its application to robot shaping.
\newblock \emph{Adaptive Behaviour} 19(5): 335--358.

\bibitem[{LeCun et~al.(2015)LeCun, Bengio and Hinton}]{LeCun2015}
LeCun Y, Bengio Y and Hinton G (2015) Deep learning.
\newblock \emph{Nature} 521(7553): 436--444.

\bibitem[{Legg and Hutter(2007)}]{Legg2007}
Legg S and Hutter M (2007) Universal intelligence: A definition of machine
  intelligence.
\newblock \emph{Minds and Machines} 17(4): 391--444.

\bibitem[{Lenz et~al.(2015)Lenz, Lee and Saxena}]{Lenz2015}
Lenz I, Lee H and Saxena A (2015) Deep learning for detecting robotic grasps.
\newblock \emph{The International Journal of Robotics Research} 34(4-5):
  705--724.

\bibitem[{Lerer et~al.(2016)Lerer, Gross and Fergus}]{Lerer2016}
Lerer A, Gross S and Fergus R (2016) Learning physical intuition of block
  towers by example.
\newblock \emph{arXiv preprint:1603.01312} .

\bibitem[{Levine and Abbeel(2014)}]{Levine2014}
Levine S and Abbeel P (2014) Learning neural network policies with guided
  policy search under unknown dynamics.
\newblock In: \emph{Advances in Neural Information Processing Systems}. pp.
  1071--1079.

\bibitem[{Levine et~al.(2015)Levine, Finn, Darrell and Abbeel}]{Levine2015}
Levine S, Finn C, Darrell T and Abbeel P (2015) End-to-end training of deep
  visuomotor policies.
\newblock \emph{arXiv preprint:1504.00702} .

\bibitem[{Levine et~al.(2016)Levine, Pastor, Krizhevsky and
  Quillen}]{Levine2016}
Levine S, Pastor P, Krizhevsky A and Quillen D (2016) Learning hand-eye
  coordination for robotic grasping with deep learning and large-scale data
  collection.
\newblock \emph{arXiv preprint:1603.02199} .

\bibitem[{Libertus and Needham(2010)}]{Libertus2010}
Libertus K and Needham A (2010) Teach to reach: The effects of active vs.
  passive reaching experiences on action and perception.
\newblock \emph{Vision research} 50(24): 2750--2757.

\bibitem[{Lillicrap et~al.(2015)Lillicrap, Hunt, Pritzel, Heess, Erez, Tassa,
  Silver and Wierstra}]{Lillicrap2015}
Lillicrap TP, Hunt JJ, Pritzel A, Heess N, Erez T, Tassa Y, Silver D and
  Wierstra D (2015) Continuous control with deep reinforcement learning.
\newblock \emph{arXiv preprint:1509.02971} .

\bibitem[{Lin(1993)}]{Lin1993}
Lin LJ (1993) Reinforcement learning for robots using neural networks.
\newblock Technical report, DTIC Document.

\bibitem[{Lin and Mitchell(1992)}]{Lin1992}
Lin LJ and Mitchell TM (1992) Memory approaches to reinforcement learning in
  non-markovian domains.
\newblock Technical report, Department of Computer Science. Carnegie Mellon
  University.

\bibitem[{Lipton et~al.(2016)Lipton, Gao, Li, Li, Ahmed and Deng}]{Lipton2016}
Lipton ZC, Gao J, Li L, Li X, Ahmed F and Deng L (2016) Efficient exploration
  for dialog policy learning with deep bbq networks \& replay buffer spiking.
\newblock \emph{arXiv preprint:1608.05081} .

\bibitem[{Lyapunov(1992)}]{Lyapunov1992}
Lyapunov AM (1992) The general problem of the stability of motion.
\newblock \emph{International Journal of Control} 55(3): 531--534.

\bibitem[{Madai-Tahy et~al.(2016)Madai-Tahy, Otte, Hanten and Zell}]{Madai2016}
Madai-Tahy L, Otte S, Hanten R and Zell A (2016) Revisiting deep convolutional
  neural networks for rgb-d based object recognition.
\newblock In: \emph{International Conference on Artificial Neural Networks}.
  pp. 29--37.

\bibitem[{Mahendran and Vedaldi(2016)}]{Mahendran2016}
Mahendran A and Vedaldi A (2016) Visualizing deep convolutional neural networks
  using natural pre-images.
\newblock \emph{International Journal of Computer Vision} : 1--23.

\bibitem[{Maitin-Shepard et~al.(2010)Maitin-Shepard, Cusumano-Towner, Lei and
  Abbeel}]{Maitin2010}
Maitin-Shepard J, Cusumano-Towner M, Lei J and Abbeel P (2010) Cloth grasp
  point detection based on multiple-view geometric cues with application to
  robotic towel folding.
\newblock In: \emph{IEEE International Conference on Robotics and Automation}.

\bibitem[{Malmir et~al.(2015)Malmir, Sikka, Forster, Fasel, Movellan and
  Cottrell}]{Malmir2015}
Malmir M, Sikka K, Forster D, Fasel I, Movellan JR and Cottrell GW (2015) Deep
  active object recognition by joint label and action prediction.
\newblock \emph{arXiv preprint:1512.05484} .

\bibitem[{Malmir et~al.(2016)Malmir, Sikka, Forster, Movellan and
  Cottrell}]{Malmir2016}
Malmir M, Sikka K, Forster D, Movellan J and Cottrell GW (2016) Deep q-learning
  for active recognition of germs: Baseline performance on a standardized
  dataset for active learning.
\newblock In: \emph{Proceedings of the British Machine Vision Conference}.

\bibitem[{Mariet and Sra(2015)}]{Mariet2015}
Mariet Z and Sra S (2015) Diversity networks.
\newblock \emph{arXiv preprint:1511.05077} .

\bibitem[{Masson and Konidaris(2015)}]{Masson2015}
Masson W and Konidaris G (2015) Reinforcement learning with parameterized
  actions.
\newblock \emph{arXiv preprint:1509.01644} .

\bibitem[{Mathieu et~al.(2015)Mathieu, Couprie and LeCun}]{Mathieu2015}
Mathieu M, Couprie C and LeCun Y (2015) Deep multi-scale video prediction
  beyond mean square error.
\newblock \emph{arXiv preprint:1511.05440} .

\bibitem[{Meeden et~al.(1993)Meeden, McGraw and Blank}]{Meeden1993}
Meeden L, McGraw G and Blank D (1993) Emergent control and planning in an
  autonomous vehicle.
\newblock In: \emph{Proceedings of the fifteenth annual meeting of the
  cognitive science society}. pp. 735--740.

\bibitem[{Mnih et~al.(2016)Mnih, Agapiou, Osindero, Graves, Vinyals,
  Kavukcuoglu et~al.}]{Mnih2016}
Mnih V, Agapiou J, Osindero S, Graves A, Vinyals O, Kavukcuoglu K et~al. (2016)
  Strategic attentive writer for learning macro-actions.
\newblock \emph{arXiv preprint:1606.04695} .

\bibitem[{Mnih et~al.(2013)Mnih, Kavukcuoglu, Silver, Graves, Antonoglou,
  Wierstra and Riedmiller}]{Mnih2013}
Mnih V, Kavukcuoglu K, Silver D, Graves A, Antonoglou I, Wierstra D and
  Riedmiller M (2013) Playing atari with deep reinforcement learning.
\newblock In: \emph{NIPS Deep Learning Workshop}.

\bibitem[{Mnih et~al.(2015)Mnih, Kavukcuoglu, Silver, Rusu, Veness, Bellemare,
  Graves, Riedmiller, Fidjeland, Ostrovski et~al.}]{Mnih2015}
Mnih V, Kavukcuoglu K, Silver D, Rusu AA, Veness J, Bellemare MG, Graves A,
  Riedmiller M, Fidjeland AK, Ostrovski G et~al. (2015) Human-level control
  through deep reinforcement learning.
\newblock \emph{Nature} 518(7540): 529--533.

\bibitem[{Montgomery et~al.(2016)Montgomery, Ajay, Finn, Abbeel and
  Levine}]{Montgomery2016b}
Montgomery W, Ajay A, Finn C, Abbeel P and Levine S (2016) Reset-free guided
  policy search: Efficient deep reinforcement learning with stochastic initial
  states.
\newblock \emph{arXiv preprint:1610.01112} .

\bibitem[{Montgomery and Levine(2016)}]{Montgomery2016}
Montgomery W and Levine S (2016) Guided policy search as approximate mirror
  descent.
\newblock \emph{arXiv preprint:1607.04614} .

\bibitem[{Moore and Atkeson(1993)}]{Moore1993}
Moore AW and Atkeson CG (1993) Prioritized sweeping: Reinforcement learning
  with less data and less time.
\newblock \emph{Machine Learning} 13(1): 103--130.

\bibitem[{Mottaghi et~al.(2016)Mottaghi, Rastegari, Gupta and
  Farhadi}]{Mottaghi2016}
Mottaghi R, Rastegari M, Gupta A and Farhadi A (2016) " what happens if..."
  learning to predict the effect of forces in images.
\newblock \emph{arXiv preprint:1603.05600} .

\bibitem[{Mugan and Kuipers(2012)}]{Mugan2012}
Mugan J and Kuipers B (2012) Autonomous learning of high-level states and
  actions in continuous environments.
\newblock \emph{Transactions on Autonomous Mental Development} 4(1): 70--86.

\bibitem[{Nakamura(1990)}]{Nakamura1990}
Nakamura Y (1990) \emph{Advanced robotics: redundancy and optimization}.
\newblock Addison-Wesley Longman Publishing Co., Inc.

\bibitem[{Ng and Lin(2016)}]{Ng2016}
Ng A and Lin Y (2016) Self-driving cars won’t work until we change our
  roads---and attitudes.

\bibitem[{Nguyen et~al.(2015)Nguyen, Yosinski and Clune}]{Nguyen2015}
Nguyen A, Yosinski J and Clune J (2015) Deep neural networks are easily fooled:
  High confidence predictions for unrecognizable images.
\newblock In: \emph{IEEE Conference on Computer Vision and Pattern Recognition
  (CVPR)}.

\bibitem[{Nicolai et~al.(2016)Nicolai, Skeele, Eriksen and
  Hollinger}]{Nicolai-RSS2016}
Nicolai A, Skeele R, Eriksen C and Hollinger GA (2016) Deep learning for laser
  based odometry estimation .

\bibitem[{Noda et~al.(2013)Noda, Arie, Suga and Ogata}]{Noda2013}
Noda K, Arie H, Suga Y and Ogata T (2013) Multimodal integration learning of
  object manipulation behaviors using deep neural networks.
\newblock In: \emph{IEEE/RSJ International Conference on Intelligent Robots and
  Systems}. pp. 1728--1733.

\bibitem[{Noh et~al.(2015)Noh, Hong and Han}]{Noh2015}
Noh H, Hong S and Han B (2015) Learning deconvolution network for semantic
  segmentation.
\newblock In: \emph{Proceedings of the IEEE International Conference on
  Computer Vision}. pp. 1520--1528.

\bibitem[{Oh et~al.(2015)Oh, Guo, Lee, Lewis and Singh}]{Oh2015}
Oh J, Guo X, Lee H, Lewis RL and Singh S (2015) Action-conditional video
  prediction using deep networks in atari games.
\newblock In: \emph{Advances in Neural Information Processing Systems}. pp.
  2863--2871.

\bibitem[{Ondruska et~al.(2016)Ondruska, Dequaire, Wang and
  Posner}]{Ondruska-RSS2016}
Ondruska P, Dequaire J, Wang DZ and Posner I (2016) End-to-end tracking and
  semantic segmentation using recurrent neural networks .

\bibitem[{Ondruska and Posner(2016)}]{Ondruska2016}
Ondruska P and Posner I (2016) Deep tracking: Seeing beyond seeing using
  recurrent neural networks.
\newblock \emph{arXiv preprint:1602.00991} .

\bibitem[{Osband et~al.(2016)Osband, Blundell, Pritzel and
  Van~Roy}]{Osband2016}
Osband I, Blundell C, Pritzel A and Van~Roy B (2016) Deep exploration via
  bootstrapped dqn.
\newblock \emph{arXiv preprint:1602.04621} .

\bibitem[{Oudeyer et~al.(2013)Oudeyer, Baranes and Kaplan}]{Oudeyer2013}
Oudeyer PY, Baranes A and Kaplan F (2013) Intrinsically motivated learning of
  real-world sensorimotor skills with developmental constraints.
\newblock In: Baldassarre G and Mirolli M (eds.) \emph{Intrinsically Motivated
  Learning in Natural and Artificial Systems}. Springer.

\bibitem[{Oudeyer et~al.(2007)Oudeyer, Kaplan and Hafner}]{Oudeyer2007}
Oudeyer PY, Kaplan F and Hafner V (2007) Intrinsic motivation systems for
  autonomous mental development.
\newblock \emph{IEEE Transactions on Evolutionary Computation} 11(2): 265--286.

\bibitem[{Peng and Williams(1993)}]{Peng1993}
Peng J and Williams RJ (1993) Efficient learning and planning within the dyna
  framework.
\newblock \emph{Adaptive Behavior} 1(4): 437--454.

\bibitem[{Peters et~al.(2010)Peters, M{\"u}lling and Altun}]{Peters2010}
Peters J, M{\"u}lling K and Altun Y (2010) Relative entropy policy search.
\newblock In: \emph{Association for the Advancement in Artificial
  Intelligence}.

\bibitem[{Piaget(1953)}]{Piaget1953}
Piaget J (1953) The origin of intelligence in the child.

\bibitem[{Piaget(1954)}]{Piaget1954}
Piaget J (1954) \emph{The construction of reality in the child}.
\newblock Routledge.

\bibitem[{Piaget and Cook(1952)}]{Piaget1952}
Piaget J and Cook M (1952) \emph{The origins of intelligence in children},
  volume~8.
\newblock International Universities Press New York.

\bibitem[{Pinheiro et~al.(2015)Pinheiro, Collobert and Dollar}]{Pinheiro2015}
Pinheiro PO, Collobert R and Dollar P (2015) Learning to segment object
  candidates.
\newblock In: \emph{Advances in Neural Information Processing Systems}. pp.
  1990--1998.

\bibitem[{Pinheiro et~al.(2016)Pinheiro, Lin, Collobert and
  Doll{\'a}r}]{Pinheiro2016}
Pinheiro PO, Lin TY, Collobert R and Doll{\'a}r P (2016) Learning to refine
  object segments.
\newblock \emph{arXiv preprint:1603.08695} .

\bibitem[{Pinto et~al.(2016)Pinto, Davidson and Gupta}]{Pinto2016b}
Pinto L, Davidson J and Gupta A (2016) Supervision via competition: Robot
  adversaries for learning tasks.
\newblock \emph{arXiv preprint:1610.01685} .

\bibitem[{Pinto and Gupta(2015)}]{Pinto2015}
Pinto L and Gupta A (2015) Supersizing self-supervision: Learning to grasp from
  50k tries and 700 robot hours.
\newblock \emph{arXiv preprint:1509.06825} .

\bibitem[{Pinto and Gupta(2016)}]{Pinto2016}
Pinto L and Gupta A (2016) Learning to push by grasping: Using multiple tasks
  for effective learning.
\newblock \emph{arXiv preprint:1609.09025} .

\bibitem[{Platt(2006)}]{Platt2006}
Platt R (2006) \emph{Learning and Generalizing Control-Based Grasping and
  Manipulation Skills}.
\newblock PhD Thesis, University of Massachusetts Amherst.

\bibitem[{Platt et~al.(2005)Platt, Fagg and Grupen}]{Platt-Humanoids05}
Platt R, Fagg A and Grupen R (2005) Reusing schematic grasping policies.
\newblock In: \emph{Proceedings of the IEEE-RAS International Conference on
  Humanoid Robots}. Tsukuba, Japan.

\bibitem[{Platt et~al.(2010)Platt, Fagg and Grupen}]{Platt-TRO10}
Platt R, Fagg A and Grupen R (2010) Null-space grasp control: Theory and
  experiments.
\newblock \emph{Robotics, IEEE Transactions on} 26(2): 282 --295.
\newblock \doi{10.1109/TRO.2010.2042754}.

\bibitem[{Platt et~al.(2002)Platt, Fagg and Grupen}]{Platt_IROS02}
Platt R, Fagg AH and Grupen R (2002) Nullspace composition of control laws for
  grasping.
\newblock In: \emph{International Conference on Intelligent Robots and Systems
  (IROS)}. Laussane, Switzerland.

\bibitem[{Platt et~al.(2004)Platt, Fagg and Grupen}]{Platt_ICRA04}
Platt R, Fagg AH and Grupen R (2004) Manipulation gaits: Sequences of grasp
  control tasks.
\newblock In: \emph{Proceedings of the 2004 IEEE Conference on Robotics and
  Automation}. New Orleans, LA.

\bibitem[{Platt et~al.(2006)Platt, Grupen and Fagg}]{Platt-ICDL06}
Platt R, Grupen R and Fagg A (2006) Improving grasp skills using schema
  structured learning.
\newblock In: \emph{Proceedings of International Conference on Developmental
  Learning}. Bloomington, Indiana.

\bibitem[{Purves et~al.(2001)Purves, Augustine, Fitzpatrick, Katz, LaMantia,
  McNamara and Williams}]{Purves2001}
Purves D, Augustine GJ, Fitzpatrick D, Katz LC, LaMantia AS, McNamara JO and
  Williams SM (2001) \emph{The Spinal Cord Circuitry Underlying Muscle Stretch
  Reflexes}.
\newblock Sinauer Associates.

\bibitem[{Radford et~al.(2015)Radford, Metz and Chintala}]{Radford2015}
Radford A, Metz L and Chintala S (2015) Unsupervised representation learning
  with deep convolutional generative adversarial networks.
\newblock \emph{arXiv preprint:1511.06434} .

\bibitem[{Reed(1996)}]{Reed96}
Reed E (1996) \emph{Encountering the World}.
\newblock New York, NY: Oxford University Press.

\bibitem[{Ricard and Kolitz(2003)}]{Ricard2003}
Ricard M and Kolitz S (2003) The adept framework for intelligent autonomy.
\newblock Technical report, DTIC Document.

\bibitem[{Rohanimanesh et~al.(2004)Rohanimanesh, Platt, Mahadevan and
  Grupen}]{Rohanimanesh2004}
Rohanimanesh K, Platt R, Mahadevan S and Grupen R (2004) Coarticulation in
  markov decision processes.
\newblock In: \emph{Advances in Neural Information Processing Systems}. pp.
  1137--1144.

\bibitem[{Ruiken et~al.(2013)Ruiken, Lanighan and Grupen}]{Ruiken2013}
Ruiken D, Lanighan MW and Grupen RA (2013) Postural modes and control for
  dexterous mobile manipulation: the umass ubot concept.
\newblock In: \emph{IEEE-RAS International Conference on Humanoid Robots}.

\bibitem[{Ruiken et~al.(2016{\natexlab{a}})Ruiken, Liu, Takahashi and
  Grupen}]{Ruiken2016b}
Ruiken D, Liu TQ, Takahashi T and Grupen RA (2016{\natexlab{a}}) Reconfigurable
  tasks in belief-space planning.
\newblock In: \emph{Integrating Multiple Knowledge Representation and Reasoning
  Techniques in Robotics Workshop at IEEE/RSJ International Conference on
  Intelligent Robots and Systems (IROS)}.

\bibitem[{Ruiken et~al.(2016{\natexlab{b}})Ruiken, Wong, Liu, Hebert,
  Takahashi, Lanighan and Grupen}]{Ruiken2016}
Ruiken D, Wong JM, Liu TQ, Hebert M, Takahashi T, Lanighan MW and Grupen RA
  (2016{\natexlab{b}}) Affordance-based active belief recognition using visual
  and manual actions.
\newblock In: \emph{IEEE/RSJ the International Conference on Intelligent Robots
  and Systems}.

\bibitem[{Rumelhart(1998)}]{Rumelhart1998}
Rumelhart DE (1998) The architecture of mind: A connectionist approach.
\newblock \emph{Mind readings} : 207--238.

\bibitem[{Rusu et~al.(2016{\natexlab{a}})Rusu, Rabinowitz, Desjardins, Soyer,
  Kirkpatrick, Kavukcuoglu, Pascanu and Hadsell}]{Rusu2016}
Rusu AA, Rabinowitz NC, Desjardins G, Soyer H, Kirkpatrick J, Kavukcuoglu K,
  Pascanu R and Hadsell R (2016{\natexlab{a}}) Progressive neural networks.
\newblock \emph{arXiv preprint:1606.04671} .

\bibitem[{Rusu et~al.(2016{\natexlab{b}})Rusu, Vecerik, Roth{\"o}rl, Heess,
  Pascanu and Hadsell}]{Rusu2016b}
Rusu AA, Vecerik M, Roth{\"o}rl T, Heess N, Pascanu R and Hadsell R
  (2016{\natexlab{b}}) Sim-to-real robot learning from pixels with progressive
  nets.
\newblock \emph{arXiv preprint:1610.04286} .

\bibitem[{Santana and Hotz(2016)}]{Santana2016}
Santana E and Hotz G (2016) Learning a driving simulator.
\newblock \emph{arXiv preprint:1608.01230} .

\bibitem[{Schaul et~al.(2015)Schaul, Quan, Antonoglou and Silver}]{Schaul2015}
Schaul T, Quan J, Antonoglou I and Silver D (2015) Prioritized experience
  replay.
\newblock \emph{arXiv preprint:1511.05952} .

\bibitem[{Schmidhuber(1991)}]{Jurgen1991}
Schmidhuber JH (1991) Reinforcement learning in markovian and non-markovian
  environments.
\newblock In: \emph{Advances in Neural Information Processing Systems}. pp.
  500--506.

\bibitem[{Schulman et~al.(2015)Schulman, Levine, Moritz, Jordan and
  Abbeel}]{Schulman2015}
Schulman J, Levine S, Moritz P, Jordan MI and Abbeel P (2015) Trust region
  policy optimization.
\newblock \emph{arXiv preprint:1502.05477} .

\bibitem[{Schwarz et~al.(2015)Schwarz, Schulz and Behnke}]{Schwarz2015}
Schwarz M, Schulz H and Behnke S (2015) Rgb-d object recognition and pose
  estimation based on pre-trained convolutional neural network features.
\newblock In: \emph{IEEE International Conference on Robotics and Automation}.

\bibitem[{Sen and Grupen(2014)}]{Sen2014}
Sen S and Grupen R (2014) Integrating task level planning with stochastic
  control.
\newblock Technical Report UM-CS-2014-005, University of Massachusetts Amherst.

\bibitem[{Sequeira et~al.(2014)Sequeira, Melo and Paiva}]{Sequeira2014}
Sequeira P, Melo FS and Paiva A (2014) {Learning by appraising: an
  emotion-based approach to intrinsic reward design}.
\newblock \emph{Adaptive Behavior} .

\bibitem[{Sermanet et~al.(2013)Sermanet, Kavukcuoglu, Chintala and
  LeCun}]{Sermanet2013}
Sermanet P, Kavukcuoglu K, Chintala S and LeCun Y (2013) Pedestrian detection
  with unsupervised multi-stage feature learning.
\newblock In: \emph{Proceedings of the IEEE Conference on Computer Vision and
  Pattern Recognition}. pp. 3626--3633.

\bibitem[{Sigaud and Droniou(2016)}]{Sigaud2016}
Sigaud O and Droniou A (2016) Towards deep developmental learning.
\newblock \emph{IEEE Transactions on Cognitive and Developmental Systems} 8(2):
  99--114.

\bibitem[{Silver et~al.(2016)Silver, Huang, Maddison, Guez, Sifre, van~den
  Driessche, Schrittwieser, Antonoglou, Panneershelvam, Lanctot
  et~al.}]{Silver2016}
Silver D, Huang A, Maddison CJ, Guez A, Sifre L, van~den Driessche G,
  Schrittwieser J, Antonoglou I, Panneershelvam V, Lanctot M et~al. (2016)
  Mastering the game of go with deep neural networks and tree search.
\newblock \emph{Nature} 529(7587): 484--489.

\bibitem[{Silver et~al.(2014)Silver, Lever, Heess, Degris, Wierstra and
  Riedmiller}]{Silver2014}
Silver D, Lever G, Heess N, Degris T, Wierstra D and Riedmiller MA (2014)
  Deterministic policy gradient algorithms.
\newblock In: \emph{Proceedings of the 31th International Conference on Machine
  Learning}. pp. 387--395.

\bibitem[{Silver et~al.(2013)Silver, Yang and Li}]{Silver2013}
Silver DL, Yang Q and Li L (2013) Lifelong machine learning systems: Beyond
  learning algorithms.
\newblock In: \emph{AAAI Spring Symposium: Lifelong Machine Learning}. pp.
  49--55.

\bibitem[{Socher et~al.(2011)Socher, Lin, Manning and Ng}]{Socher2011}
Socher R, Lin CC, Manning C and Ng AY (2011) Parsing natural scenes and natural
  language with recursive neural networks.
\newblock In: \emph{Proceedings of the 28th international conference on machine
  learning (ICML-11)}. pp. 129--136.

\bibitem[{Song and Xiao(2015)}]{Song2015}
Song S and Xiao J (2015) Deep sliding shapes for amodal 3d object detection in
  rgb-d images.
\newblock \emph{arXiv preprint:1511.02300} .

\bibitem[{Stolle and Precup(2002)}]{Stolle2002}
Stolle M and Precup D (2002) Learning options in reinforcement learning.
\newblock In: \emph{International Symposium on Abstraction, Reformulation, and
  Approximation}. Springer, pp. 212--223.

\bibitem[{Sutton and Barto(1998)}]{SuttonBarto98}
Sutton R and Barto A (1998) \emph{Reinforcement Learning: An Introduction}.
\newblock Cambridge, MA: MIT Press.

\bibitem[{Sutton et~al.(1999)Sutton, Precup and Singh}]{Sutton-AIJ99}
Sutton R, Precup D and Singh S (1999) Between mdps and semi-mdps: A framework
  for temporal abstraction in reinforcement learning.
\newblock \emph{Artificial Intelligence} 112: 181--211.

\bibitem[{Sutton(1990)}]{Sutton1990}
Sutton RS (1990) Integrated architectures for learning, planning, and reacting
  based on approximating dynamic programming.
\newblock In: \emph{Proceedings of the seventh international conference on
  machine learning}. pp. 216--224.

\bibitem[{Sutton(1991)}]{Sutton1991}
Sutton RS (1991) Planning by incremental dynamic programming.
\newblock In: \emph{Proceedings of the Eighth International Workshop on Machine
  Learning}. pp. 353--357.

\bibitem[{Szegedy et~al.(2015)Szegedy, Liu, Jia, Sermanet, Reed, Anguelov,
  Erhan, Vanhoucke and Rabinovich}]{Szegedy2015}
Szegedy C, Liu W, Jia Y, Sermanet P, Reed S, Anguelov D, Erhan D, Vanhoucke V
  and Rabinovich A (2015) Going deeper with convolutions.
\newblock In: \emph{Proceedings of the IEEE Conference on Computer Vision and
  Pattern Recognition}. pp. 1--9.

\bibitem[{Szegedy et~al.(2013)Szegedy, Zaremba, Sutskever, Bruna, Erhan,
  Goodfellow and Fergus}]{Szegedy2013}
Szegedy C, Zaremba W, Sutskever I, Bruna J, Erhan D, Goodfellow I and Fergus R
  (2013) Intriguing properties of neural networks.
\newblock \emph{arXiv preprint:1312.6199} .

\bibitem[{Tai and Liu(2016)}]{Tai2016}
Tai L and Liu M (2016) Towards cognitive exploration through deep reinforcement
  learning for mobile robots.
\newblock \emph{arXiv preprint:1610.01733} .

\bibitem[{Takahashi et~al.(2017)Takahashi, Wong and Grupen}]{Takahashi2017}
Takahashi T, Wong JM and Grupen RA (2017) Self-supervised deep sensorimotor
  learning using closed loop motion primitives.
\newblock In: \emph{Submitted to IEEE International Conference on Robotics and
  Automation}.

\bibitem[{Tamar et~al.(2016)Tamar, Levine and Abbeel}]{Tamar2016}
Tamar A, Levine S and Abbeel P (2016) Value iteration networks.
\newblock \emph{arXiv preprint:1602.02867} .

\bibitem[{Tanay and Griffin(2016)}]{Tanay2016}
Tanay T and Griffin L (2016) A boundary tilting persepective on the phenomenon
  of adversarial examples.
\newblock \emph{arXiv preprint:1608.07690} .

\bibitem[{Tesauro(1992)}]{Tesauro1992}
Tesauro G (1992) Practical issues in temporal difference learning.
\newblock In: \emph{Reinforcement Learning}. pp. 33--53.

\bibitem[{Tesauro(1995{\natexlab{a}})}]{Tesauro1995}
Tesauro G (1995{\natexlab{a}}) Td-gammon: A self-teaching backgammon program.
\newblock In: \emph{Applications of Neural Networks}. Springer, pp. 267--285.

\bibitem[{Tesauro(1995{\natexlab{b}})}]{Tesauro1995b}
Tesauro G (1995{\natexlab{b}}) Temporal difference learning and td-gammon.
\newblock \emph{Communications of the ACM} 38(3): 58--68.

\bibitem[{Thelen and Smith(1996)}]{Thelen1996}
Thelen E and Smith L (1996) \emph{A Dynamic Systems Approach to the Development
  of Cognition and Action}.
\newblock A Bradford book. MIT Press.
\newblock ISBN 9780262700597.

\bibitem[{Thomas(2015)}]{Thomas-Thesis2015}
Thomas PS (2015) Safe reinforcement learning.
\newblock Technical report, Univeristy of Massachusetts Amherst.

\bibitem[{Thomas et~al.(2015)Thomas, Theocharous and Ghavamzadeh}]{Thomas2015}
Thomas PS, Theocharous G and Ghavamzadeh M (2015) High-confidence off-policy
  evaluation.
\newblock In: \emph{Association for the Advancement of Artificial
  Intelligence}. pp. 3000--3006.

\bibitem[{Toshev and Szegedy(2014)}]{Toshev2014}
Toshev A and Szegedy C (2014) Deeppose: Human pose estimation via deep neural
  networks.
\newblock In: \emph{Proceedings of the IEEE Conference on Computer Vision and
  Pattern Recognition}. pp. 1653--1660.

\bibitem[{Turvey(1992)}]{Turvey92}
Turvey M (1992) Affordances and prospective control: An outline of the
  ontology.
\newblock \emph{Ecological Psychology} 4: 173--187.

\bibitem[{Tzeng et~al.(2015)Tzeng, Devin, Hoffman, Finn, Peng, Levine, Saenko
  and Darrell}]{Tzeng2015}
Tzeng E, Devin C, Hoffman J, Finn C, Peng X, Levine S, Saenko K and Darrell T
  (2015) Towards adapting deep visuomotor representations from simulated to
  real environments.
\newblock \emph{arXiv preprint:1511.07111} .

\bibitem[{Utgoff and Stracuzzi(2002)}]{Utgoff2002}
Utgoff PE and Stracuzzi DJ (2002) Many-layered learning.
\newblock \emph{Neural Computation} 14(10): 2497--2529.

\bibitem[{Van~Hasselt et~al.(2015)Van~Hasselt, Guez and
  Silver}]{vanHasselt2015}
Van~Hasselt H, Guez A and Silver D (2015) Deep reinforcement learning with
  double q-learning.
\newblock \emph{arXiv preprint:1509.06461} .

\bibitem[{van Hoof et~al.(2016)van Hoof, Chen, Karl, van~der Smagt and
  Peters}]{vanHoof2016}
van Hoof H, Chen N, Karl M, van~der Smagt P and Peters J (2016) Stable
  reinforcement learning with autoencoders for tactile and visual data.
\newblock International Conference on Intelligent Robots and Systems.

\bibitem[{Varley et~al.(2016)Varley, DeChant, Richardson, Nair, Ruales and
  Allen}]{Varley2016}
Varley J, DeChant C, Richardson A, Nair A, Ruales J and Allen P (2016) Shape
  completion enabled robotic grasping.
\newblock \emph{arXiv preprint:1609.08546} .

\bibitem[{Wang et~al.(2016{\natexlab{a}})Wang, Guo, Ororbia, Alexander, Xing,
  Lin, Giles, Liu, Liu and Xiong}]{Wang2016b}
Wang Q, Guo W, Ororbia I, Alexander G, Xing X, Lin L, Giles CL, Liu X, Liu P
  and Xiong G (2016{\natexlab{a}}) Using non-invertible data transformations to
  build adversary-resistant deep neural networks.
\newblock \emph{arXiv preprint:1610.01934} .

\bibitem[{Wang and Chirikjian(2000)}]{Wang2000}
Wang Y and Chirikjian GS (2000) A new potential field method for robot path
  planning 2: 977--982.

\bibitem[{Wang et~al.(2015)Wang, de~Freitas and Lanctot}]{Wang2015}
Wang Z, de~Freitas N and Lanctot M (2015) Dueling network architectures for
  deep reinforcement learning.
\newblock \emph{arXiv preprint:1511.06581} .

\bibitem[{Wang et~al.(2016{\natexlab{b}})Wang, Li, Wang and Liu}]{Wang2016}
Wang Z, Li Z, Wang B and Liu H (2016{\natexlab{b}}) Robot grasp detection using
  multimodal deep convolutional neural networks.
\newblock \emph{Advances in Mechanical Engineering} 8(9).

\bibitem[{Wawrzy{\'n}ski and Tanwani(2013)}]{Wawrzynski2013}
Wawrzy{\'n}ski P and Tanwani AK (2013) Autonomous reinforcement learning with
  experience replay.
\newblock \emph{Neural Networks} 41: 156--167.

\bibitem[{Wilkinson and Takahashi(2015)}]{Wilkinson2015}
Wilkinson E and Takahashi T (2015) Efficient aspect object models using
  pre-trained convolutional neural networks.
\newblock In: \emph{IEEE-RAS 15th International Conference on Humanoid Robots}.
  pp. 284--289.

\bibitem[{Wong and Grupen(2016)}]{Wong-ICDL2016}
Wong JM and Grupen RA (2016) Intrinsically motivated multimodal structure
  learning.
\newblock In: \emph{IEEE International Conference on Development and Learning
  and on Epigenetic Robotics}.

\bibitem[{Wong et~al.(2016)Wong, Takahashi and Grupen}]{Wong-RSS2016}
Wong JM, Takahashi T and Grupen RA (2016) Self-supervised deep visuomotor
  learning from motor unit feedback.
\newblock In: \emph{Are the Skeptics Right? Limits and Potentials of Deep
  Learning in Robotics Workshop at Robotics: Science and Systems Conference
  (RSS-WS)}.

\bibitem[{Worgotter et~al.(2015)Worgotter, Geib, Tamosiunaite, Aksoy, Piater,
  Xiong, Ude, Nemec, Kraft, Kruger, Wächter and Asfour}]{Worgotter2015}
Worgotter F, Geib C, Tamosiunaite M, Aksoy EE, Piater J, Xiong H, Ude A, Nemec
  B, Kraft D, Kruger N, Wächter M and Asfour T (2015) Structural bootstrapping
  - a novel, generative mechanism for faster and more efficient acquisition of
  action-knowledge.
\newblock \emph{IEEE Transactions on Autonomous Mental Development} 7(2):
  140--154.

\bibitem[{Wray et~al.(2016)Wray, Ruiken, Grupen and Zilberstein}]{Wray2016}
Wray KH, Ruiken D, Grupen RA and Zilberstein S (2016) Log-space harmonic
  function path planning.
\newblock In: \emph{Twenty-Ninth International Conference on Intelligent Robots
  and Systems}.

\bibitem[{Yi et~al.(2015)Yi, McGill, Vadakedathu, He, Ha, Han, Song, Rouleau,
  Zhang, Hong et~al.}]{Yi2015}
Yi SJ, McGill SG, Vadakedathu L, He Q, Ha I, Han J, Song H, Rouleau M, Zhang
  BT, Hong D et~al. (2015) Team thor's entry in the darpa robotics challenge
  trials 2013.
\newblock \emph{Journal of Field Robotics} 32(3): 315--335.

\bibitem[{Zelazo(1983)}]{Zelazo1983}
Zelazo PR (1983) The development of walking: new findings and old assumptions.
\newblock \emph{Journal of Motor Behavior} 15(2): 99--137.

\bibitem[{Zhai et~al.(2016)Zhai, Liu, Zhang, Zhong, Zhu, Zhang and
  Sun}]{Zhai2016}
Zhai J, Liu Q, Zhang Z, Zhong S, Zhu H, Zhang P and Sun C (2016) Deep
  q-learning with prioritized sampling.
\newblock In: \emph{International Conference on Neural Information Processing}.
  Springer, pp. 13--22.

\bibitem[{Zhang et~al.(2015)Zhang, Leitner, Milford, Upcroft and
  Corke}]{Zhang2015}
Zhang F, Leitner J, Milford M, Upcroft B and Corke P (2015) Towards
  vision-based deep reinforcement learning for robotic motion control.
\newblock \emph{arXiv preprint:1511.03791} .

\bibitem[{Zheng et~al.(2015)Zheng, Jayasumana, Romera-Paredes, Vineet, Su, Du,
  Huang and Torr}]{Zheng2015}
Zheng S, Jayasumana S, Romera-Paredes B, Vineet V, Su Z, Du D, Huang C and Torr
  PH (2015) Conditional random fields as recurrent neural networks.
\newblock In: \emph{Proceedings of the IEEE International Conference on
  Computer Vision}. pp. 1529--1537.

\end{thebibliography}
	
\end{document}